\documentclass[lettersize,journal]{IEEEtran}

\usepackage[caption=false,subrefformat=parens]{subfig} 
\usepackage{cite}
\usepackage{amsmath,amssymb,amsfonts}
\usepackage[noend]{algpseudocode}
\usepackage{algorithm}
\usepackage{graphicx}
\usepackage{textcomp}
\usepackage{xcolor}
\usepackage{float}
\usepackage{picinpar}
\usepackage{babel}
\usepackage{booktabs}
\usepackage{blindtext}
\DeclareSubrefFormat{parens}{#1(#2)}

\usepackage{hyperref}
\hypersetup{hidelinks}

\usepackage{tikz}
\definecolor{lime}{HTML}{A6CE39}
\DeclareRobustCommand{\orcidicon}{%
    \begin{tikzpicture}
        \draw[lime, fill=lime] (0,0) 
        circle [radius=0.16] 
        node[white] {{\fontfamily{qag}\selectfont \tiny ID}};
        \draw[white, fill=white] (-0.0625,0.095) 
        circle [radius=0.007];    
    \end{tikzpicture}
    \hspace{-2mm}
}
\foreach \x in {A, ..., Z}{%
    \expandafter\xdef\csname orcid\x\endcsname{\noexpand\href{https://orcid.org/\csname orcidauthor\x\endcsname}{\noexpand\orcidicon}}
}

\newcommand\SPACE{\mathcal}  
\usepackage{color}

\usepackage{bbm}
\usepackage{subfig}
\usepackage{stfloats}

\usepackage{mathtools}
\usepackage{amsthm}

\newtheorem{definition}{Definition}

\begin{document}

\title{Safe Model-Based Reinforcement Learning with an Uncertainty-Aware Reachability Certificate}

\author{
    Dongjie Yu\orcidA{},
    Wenjun Zou,
    Yujie Yang,
    Haitong Ma,~\IEEEmembership{Graduate Student Member,~IEEE},\\
    Shengbo Eben Li\orcidB{},~\IEEEmembership{Senior Member,~IEEE},
    Jingliang Duan\orcidC{} and
    Jianyu Chen
    \thanks{
        \textit{(Dongjie Yu, Wenjun Zou and Yujie Yang contributed equally to this work.)}
    }
    \thanks{
        Dongjie Yu, Wenjun Zou, Yujie Yang and Shengbo Eben Li are with the School of Vehicle and Mobility, Tsinghua University, Beijing 100084, China (email: \{ydj20,zouwj20,yangyj21\}@mails.tsinghua.edu.cn, lishbo@mail.tsinghua.edu.cn).
    }
    \thanks{
        Haitong Ma is with John A. Paulson School of Engineering and Applied Sciences, Harvard University, Cambridge, Massachusetts 02138, USA (email:haitongma@g.harvard.edu).
    }%
    \thanks{
        Jingliang Duan is with the School of Mechanical Engineering, University of Science and Technology Beijing, Beijing 100083, China (email: duanjl@ustb.edu.cn).
    }%
    \thanks{
        Jianyu Chen is with the Institute for Interdisciplinary Information Sciences, Tsinghua University, Beijing 100084, China (email: jianyuchen@tsinghua.edu.cn).
    }%
}

\markboth{Preprint. Copyright 2022 by the author(s).}{Yu \MakeLowercase{\textit{et al.}}: Safe Model-Based Reinforcement Learning with an Uncertainty-Aware Reachability Certificate}


\maketitle

\begin{abstract}
Safe reinforcement learning (RL) that solves constraint-satisfactory policies provides a promising way to the broader safety-critical applications of RL in real-world problems such as robotics. Among all safe RL approaches, model-based methods reduce training time violations further due to their high sample efficiency. However, lacking safety robustness against the model uncertainties remains an issue in safe model-based RL, especially in training time safety. In this paper, we propose a distributional reachability certificate (DRC) and its Bellman equation to address model uncertainties and characterize robust persistently safe states. Furthermore, we build a safe RL framework to resolve constraints required by the DRC and its corresponding shield policy. We also devise a line search method to maintain safety and reach higher returns simultaneously while leveraging the shield policy. Comprehensive experiments on classical benchmarks such as constrained tracking and navigation indicate that the proposed algorithm achieves comparable returns with much fewer constraint violations during training.
\end{abstract}

\def\abstractname{Note to Practitioners}
\begin{abstract}
Although it has been proven that RL can be applied in complex robotics control tasks, the training process of an RL control policy induces frequent failures because the agent needs to learn safety through constraint violations. This issue hinders the promotion of RL because a large amount of failure of robots is too expensive to afford. 
This paper aims to reduce the training-time violations of RL-based control methods, enabling RL to be leveraged in a broader application area. To achieve the goal, we first introduce a safety quantity describing the distribution of potential constraint violations in the long term. By imposing constraints on the quantile of the safety distribution, we can realize safety robust to the model uncertainty, which is necessary for real-world robot learning with environment uncertainty. Second, we further devise a shield policy aiming to minimize the constraint violation. The policy will intervene when the agent is about to violate state constraints, further enhancing exploration safety. Third, we implement a line search method to find an action pursuing near-optimal performance when fulfilling safety requirements strictly. 
Our experimental results indicate that the proposed algorithm reduces training-time violations significantly while maintaining competitive task performance. We make a step towards applying RL safely in real-world tasks.
Our future work includes conducting physical verification on real robots to evaluate the algorithm and improving safety further by starting from an initially safe control policy that comes from domain knowledge.
\end{abstract}


\section{Introduction}
\IEEEPARstart{R}{einforcement} learning (RL) has achieved success across different automated control tasks such as robotics locomotion~\cite{hafner2020Dream}, navigation~\cite{liu2021Visuomotor} and transportation management~\cite{brittain2022Scalable, yan2022Unified}. However, the trial-and-error process of RL hinders its application in more real-world tasks due to the large number of failures brought by unconstrained policies, threatening the safety of users and systems. Therefore, safe RL~\cite{altman2021Constrained} is proposed to impose constraints on agents and enhance the safety of policies both after convergence and during the training process. The training time safety issue, also called \textit{safe exploration} (i.e., reducing the number of constraint violations during learning), is thought to be challenging and significant, especially when the dynamics of the environment are unknown.
In this paper, we not only focus on the safe RL problem subject to certain constraints \textit{after convergence}, but also take a step towards \textit{safe exploration}.

Previous work formulates safe RL problems as constrained Markov decision process (CMDP)~\cite{altman2021Constrained, paternain2019Constrained}. CMDP augments MDP with an additional cost signal indicating state-action pairs that violate constraints. By setting thresholds on the costs, one gets policies with a low probability of failure. Common approaches to solve CMDPs include penalty function methods~\cite{guan2022Integrated}, Lagrange multiplier methods~\cite{duan2021Adaptive, chow2017RiskConstrained, tessler2018Reward, ray2019Benchmarking} and projection methods~\cite{yang2020ProjectionBased}. These work takes the expected cumulative costs (cost value in short) as the constrained quantity to realize safety \textit{in expectation}. In contrast, another line of work replaces the cost value with safety certificates such as energy function~\cite{ma2022Joint} or reachability certificate~\cite{yu2022Reachabilitya} to equip algorithms with \textit{persistent} safety constraint satisfaction in each state. However, all these methods utilize model-free RL (MFRL) algorithms whose low sample efficiency leads to significantly more interactions with environments, and more violations inevitably happen during learning.

An alternative way to improve training time safety is to use model-based RL (MBRL), i.e., performing policy updates with model-generated virtual data~\cite{chua2018Deep, luo2018Algorithmic, janner2019When, hafner2020Dream}. Recent studies combine CMDP with MBRL to reduce training violations and accelerate learning~\cite{thomas2021Safe, as2022Constraineda, zanger2021Safe}. Despite their progress, MBRL is confronted with the uncertainty of the learned model. Because of the insufficient learning of dynamics, the model may predict transitions deviating from true dynamics, which makes the agent potentially take unsafe states as safe ones.

To address the uncertainty in safe MBRL, we propose \textbf{D}istributional \textbf{R}eachability \textbf{P}olicy \textbf{O}ptimization (DRPO), a Lagrangian-based MBRL approach with an uncertainty-aware reachability certificate to realize robust but non-conservative (training time) safety. Moreover, the certificate produces a shield policy, which we leverage from the perspective of the optimization problem and the perspective of policy execution, to alleviate conservativeness or potential danger caused by approximation error and insufficient learning. A detailed illustration of our DRPO framework is shown in Fig~\ref{fig.framework}. Our main contributions are as follows:
\begin{figure*}[!ht]
\centering
    \subfloat[Training: the learning scheme with an action switch]{\includegraphics[width=0.48\textwidth]{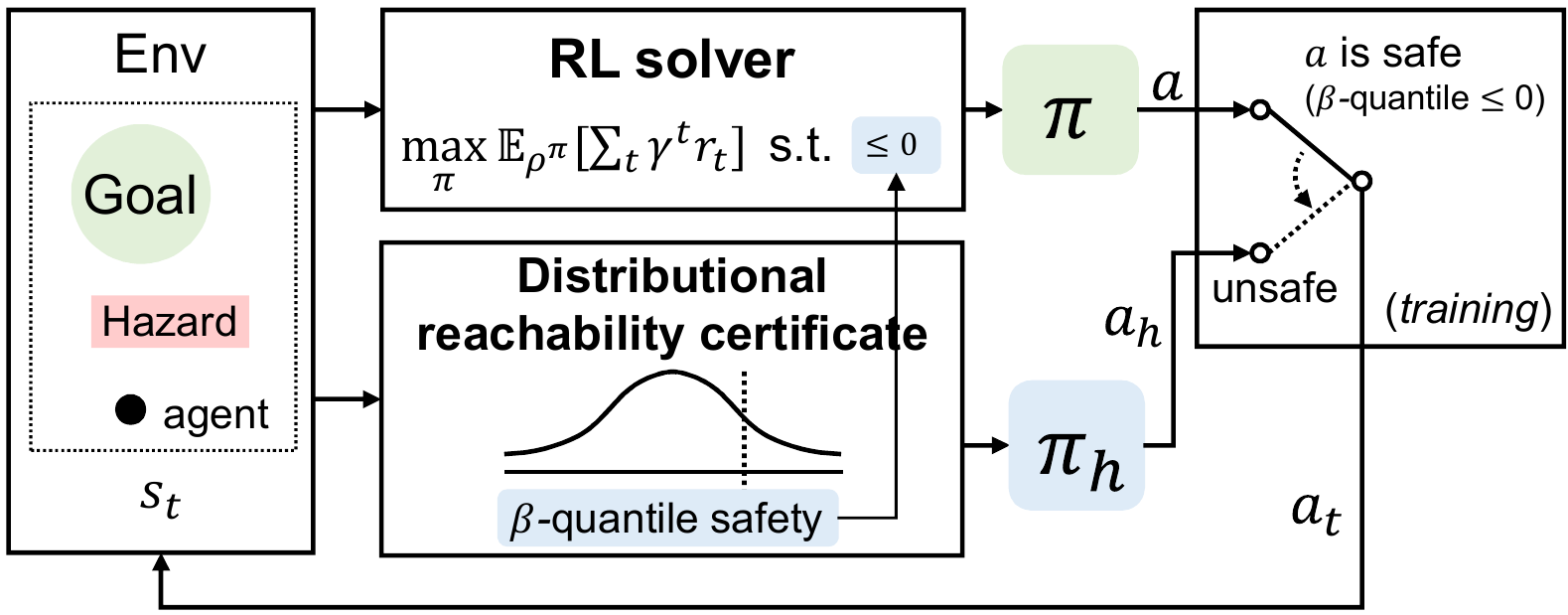}\label{fig.overview_training}} \hspace{10mm}
    \subfloat[Evaluation: safety shield with a line search method]{\includegraphics[width=0.36\textwidth]{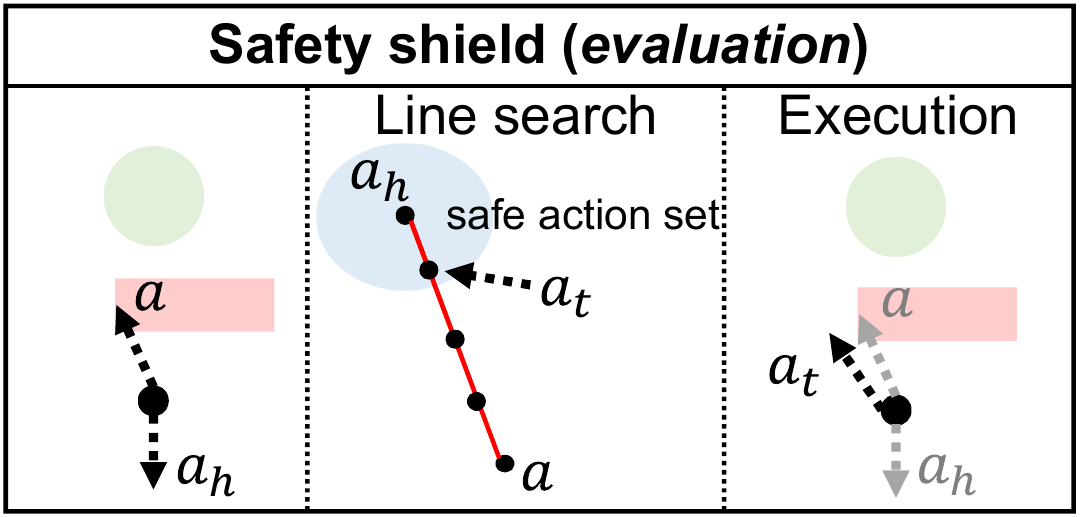}\label{fig.overview_shield}}
    \caption{\textbf{DRPO}: We illustrate DRPO intuitively with a 2D reach-avoid task where the violation corresponds to entering the hazard. We solve a constrained RL problem where the $\beta$-quantile of the distributional reachability certificate (DRC) is restricted. The RL solver produces a \textit{main policy} $\pi$ considering both cumulative rewards and safety while the DRC gives a \textit{shield policy} $\pi_h$ only considering safety. The two policies output $a$ and $a_h$ given $s_t$ at each time step, respectively. (a) Once the DRC indicates $a$ is unsafe (the $\beta$-quantile is above zero), the agent will execute $a_h$ instead during \textit{training}. (b) During \textit{evaluation}, we perform a line search between $a$ and $a_h$ in the action space to avoid conservativeness if $a$ is considered unsafe. The result $a_t$ of the search is executed.}
    \label{fig.framework}
\end{figure*}

\begin{itemize}
\setlength{\itemsep}{0pt}
\setlength{\parsep}{0pt}
\setlength{\parskip}{0pt}
    \item We propose a \textit{distributional} reachability certificate (DRC) quantifying the safety of RL policies, whose variance explicitly represents the epistemic uncertainty of the agent in terms of its potential constraint violation in the future. To this end, the pessimism for uncertainty can be considered during learning. Existing safe MBRL approaches either ignore the model uncertainty or consider the uncertain effects of the learned model on safety implicitly (e.g., sampling and estimation). Therefore, our work gives direct intuition about to which degree the distributional reachability certificate is confident about safety.

    \item We leverage the shield policy produced by the DRC in different schemes to reduce conservativeness and maintain safety. Restricting the main policy subject to the certificate of the shield policy provides a looser constraint in the optimization problem, enabling the main policy to explore freely for higher return. Furthermore, the shield serves as a backup policy during interaction because it is safety-oriented.
    We also propose a line search method for efficient and safe action during \textit{evaluation}, achieving a trade-off between performance and conservatism.

    \item Comprehensive empirical results on robotics tasks such as constrained stabilization, tracking and navigation indicate the efficacy of the proposed DRC and DRPO algorithm. DRPO reaches comparable performance w.r.t. model-free and model-based baselines while reducing training time violations significantly and converging to a safe policy.
\end{itemize}

\section{Related Work}

\subsection{Safe Model-Free RL}

Many safe model-free RL algorithms are designed under CMDP~\cite{altman2021Constrained, paternain2019Constrained}. They constrain the cost value of the policy below a given threshold. Penalty function methods augment the original policy objective function by an additional term, which is the cost value multiplied by a penalty coefficient~\cite{guan2022Integrated}. Lagrange multiplier methods convert the constrained policy optimization problem into an unconstrained problem by constructing a Lagrangian as the policy objective~\cite{duan2021Adaptive, chow2017RiskConstrained, tessler2018Reward, ray2019Benchmarking}. The policy parameters and the Lagrange multiplier are updated in an iterative manner using dual ascent. Projection methods update the policy in a two-step process~\cite{yang2020ProjectionBased}. The first step only considers the improvement of the rewards, while the second step ensures constraint satisfaction of the policy by projecting it back onto the constraint set. 

Other algorithms use safety certificates instead of cost values to constrain the policy optimization. Safety certificates theoretically guarantee that the state trajectory always stays in a subset of the state space where the safety constraint is satisfied. Energy function methods synthesize a safety index with a pre-defined functional form that assigns low energy to safe states and guarantees persistent safety of energy-dissipating policies~\cite{ma2022Joint}. Reachability certificate methods use a safety value function to describe the worst-case constraint value and constrain it below zero in policy optimization~\cite{yu2022Reachabilitya}.

There is also safe RL work adopting distributional RL to handle the stochasticity of the dynamics and enhance safety~\cite{yang2021WCSAC, yang2022Safetyconstrained, kim2022TRC} but they also suffer from conservativeness. Moreover, a common problem of these model-free algorithms is that they cannot guarantee safety during training. It is because the cost value functions and safety certificates are learned totally from interaction data, which inevitably leads to a large number of safety violations during the learning process. Moreover, the low sample efficiency of model-free RL algorithms further increases training-time violations. 

\subsection{Safe Model-Based RL}

Safe model-based RL algorithms aim at minimizing safety constraint violations during training. They learn policies on virtual data generated by environment models to improve sample efficiency and decrease the number of training-time violations in the actual environment. Thomas et al.~\cite{thomas2021Safe} avoid unsafe states by planning ahead a short time into the future. They learn an ensemble of probabilistic dynamics models for planning and heavily penalize unsafe trajectories to ensure constraint satisfaction. As et al.~\cite{as2022Constraineda} use Bayesian world models to approximate the true dynamics. They estimate optimistic upper bounds on the task objective and pessimistic upper bounds on the safety constraints with posterior sampling. The augmented Lagrangian method solves the constrained policy optimization problem based on the estimated bounds. Zanger et al.~\cite{zanger2021Safe} address the problem of accumulative model errors by adaptive resampling from an ensemble of probabilistic environment models and using dynamically limited rollout horizons.
Our method differs from these work in that we explicitly model the uncertainty of the safety value caused by the learned probabilistic model and learn a robust safe policy to address the uncertainty.

\section{Preliminary}

\subsection{Safe Model-Based RL}

\textbf{Constrained Markov decision process} is the common formulation of safe RL problems. We consider an infinite-horizon CMDP defined by the tuple $\langle \SPACE{S}, \SPACE{A}, P, r, c, \gamma\rangle$, where (1) the state space $\SPACE{S}$ and the action space $\SPACE{A}$ are bounded and possibly continuous; (2) the unknown transition dynamics $P: \SPACE{S} \times \SPACE{A}\mapsto \Delta(\SPACE{S})$ gives the distribution of the next state $P(\cdot \mid s_t, a_t)$, where $\Delta(\SPACE{S})$ is a distribution on the space $\SPACE{S}$; (3) $r: \SPACE{S}\times\SPACE{A} \mapsto \mathbb{R}$ is the reward function; (4) $c$ is called the cost signal where $c(s,a)=\mathbbm{1}_{h(s,a)>0}$, indicating we get $1$ and the episode is terminated once the desired state constraint $h(s,a)\le0$ is violated and otherwise $0$; (5) $\gamma \in [0,1)$ is the discounted factor. At each time step $t\in \mathbb{N}$, the agent observes state $s_t$ and chooses its action $a_t$ according to the policy $\pi(\cdot|s_t)\in\Pi$, where $\Pi$ is the set of all Markovian stationary polices. Then the dynamics transit to state $s_{t+1}$ (sometimes we denote it as $s'$) and send reward $r_t$, cost $c_t$, constraint violation $h_t$ to the agent. Given the initial state distribution $d_0(s)$, we have $d_{\pi}(s,a)\coloneqq \sum_{t}^{\infty} \gamma^t\mathbb{P}(s_t=s|d_0,\pi,P)\pi(a|s)$ as the state-action marginals following $\pi$. We also denote the initial state set as $\SPACE{S}_0\coloneqq\{s\mid d_0(s)>0\}$. The purpose of safe RL is to solve the following problem:
\begin{equation}
\begin{aligned}
\label{eq.srl_problem}
\max_{\pi\in\Pi} & \quad \mathbb{E}_{s\sim d_0(s),a\sim\pi(\cdot|s)} [Q^{\pi}(s,a)] \\
\mbox{s.t.} & \quad  \mathbb{E}_{s\sim d_0(s),a\sim\pi(\cdot|s)}[Q^{\pi}_c(s,a)] \le \eta,
\end{aligned}
\end{equation}
where the state-action value $Q^{\pi}(s,a)\coloneqq\mathbb{E}_{(s,a)\sim d_{\pi}}[\sum_{t} \gamma^t r_t]$ is the expected discounted cumulative rewards starting from $(s,a)$ following $\pi$; the cost value $Q_c^{\pi}(s,a)\coloneqq\mathbb{E}_{(s,a)\sim d_{\pi}}[\sum_{t} \gamma^t c_t]$ is the expected discounted violation probability and $\eta\in[0,1)$ is the cost threshold.

\textbf{Model-based RL}~\cite{chua2018Deep, janner2019When, hafner2019Learning, hafner2020Dream} replaces the unknown transition dynamics with a learned model $\hat{P}$ which is trained by minimizing $\mathbb{E}_{(s,a,s')\sim\mathcal{B}}[D(P,\hat{P})]$, where $D$ is a certain distance metric and $\mathcal{B}$ is either an offline dataset of state-action pairs or a replay buffer storing historical interactions. That is, the samples $(s,a,s')$ are generated by the learned model rather than the true dynamics. If the learned model is accurate enough, the virtual data are similar to the ones from interaction with the real environment, which guarantees the performance of MBRL and reduces exploration and potential violation in the real environment~\cite{szita2010Modelbased, luo2018Algorithmic}. Another line of MBRL is to utilize the learned model to perform planning instead of policy learning, which is also known as model predictive control (MPC)~\cite{hafner2019Learning, liu2021Constrained}.

\subsection{Reachability Certificate}
Besides the cost value, researchers are working on advanced quantities describing the safety of policies, including control barrier function (CBF)~\cite{choi2021Robust, dawson2022Safe} or barrier certificate, safe index (SI)~\cite{ma2022Joint} and reachability certificate~\cite{fisac2019bridging, hsu2021Safety, yu2022Reachabilitya}. These functions emphasize a vital property of safety-critical systems, the forward invariance property, i.e., satisfying constraints persistently. Therefore, replacing the constraints in~\eqref{eq.srl_problem} with certificates mentioned above enables the system to avoid violation at \textit{each time step} instead of \textit{in expectation}, and then brings stricter safety. Among all these certificates, the reachability certificate aims to locate as many persistently safe states as possible~\cite{fisac2015Reachavoid} and characterize the potentially safe states without handcrafted thresholds~\cite{qin2021density}. In safe RL, more accessible safe states lead to a larger workspace for the agent to explore safely and hence possibly higher returns. Thus, we adopt reachability certificate in this work as well.

Rather than cumulative costs in the cost value, reachability certificate focuses on the worst-case discounted constraint violation starting from the state-action pair $(s,a)$. We first discuss about a deterministic $P$ and $\pi$ here for simplicity and the stochastic case is illustrated in detail in Section~\ref{sec.drc}:
\begin{equation}
\label{eq.def_qh}
Q_h^\pi(s,a) \coloneqq \max\limits_{t\in \mathbb{N}} \gamma^t h(s_t,a_t)\mid s_0=s,a_0=a,\pi, P,
\end{equation}
which means after taking action $a$ at state $s$, the agent follows the policy $\pi$ and observes the worst-case discounted constraint value. Note that the discount factor $\gamma$ is for the convergence convenience and $\gamma<1$ means we pay more attention to the constraint in the near future. If (1) $ \min_{a\in\SPACE{A}}Q_h^\pi(s,a) \le 0 $, there must exist an action $a^*$ guaranteeing the safety of the agent in the infinite horizon starting from $s$; (2) $\min_{a\in\SPACE{A}}Q_h^\pi(s,a) > 0 $, the agent is doomed to violate the constraint $h(s,a) \le 0$ in the future following $\pi$. We define the persistently safe states in the former case as feasible states:

\begin{definition}[Persistently safe states]
\label{def.feasible states}
All states starting from which the constraint will not be violated if following a given policy $\pi$ are defined persistently safe feasible states. They are included in the feasible set of $\pi$
\begin{equation}
\SPACE{S}_f^\pi \coloneqq \{s\in\SPACE{S}: \min_{a\in\SPACE{A}}Q_h^\pi(s,a) \le 0\}.
\end{equation}
\end{definition}

Hence, we can characterize the persistently safe states in safe RL by computing the reachability certificate~\cite{fisac2019bridging, yu2022Reachabilitya}, which follows an equation like the Bellman equation of $Q^\pi$:
\begin{equation}
\label{eq.sbe}
Q_h^\pi(s,a)=(1-\gamma)h(s,a) +\gamma\max\left\{h(s,a), Q_h^\pi(s',\pi(s'))\right\}.
\end{equation}
Till now, we can perform dynamic programming or temporal-difference learning to calculate the reachability certificate like the computation of the Q function in conventional RL.

\subsection{Starting Problem Formulation}
Leveraging the aforementioned reachability certificate in safe RL problems~\eqref{eq.srl_problem} to replace the constraint imposed on cost value functions, we can get a more appropriate formulation characterizing the persistently safe states rather than the safe ones in expectation~\cite{fisac2019bridging, hsu2021Safety, yu2022Reachabilitya}:
\begin{equation}
\begin{aligned}
\label{eq.rcrl_problem}
\max_{\pi\in\Pi} & \quad \mathbb{E}_{s\sim d_0(s)} [Q^{\pi}(s,a)\cdot\mathbbm{1}_{s\in \SPACE{S}_f^\pi} - Q_h^\pi(s,a) \cdot \mathbbm{1}_{s\notin \SPACE{S}_f^\pi}] \\
\mbox{s.t.} & \quad Q_h^\pi(s,a) \le 0, \forall s\in \SPACE{S}_0 \cap \SPACE{S}_f^\pi
\end{aligned},
\end{equation}
where $a$ is the action taken at state $s$ without exploration noise given by policy $\pi$ (thus a deterministic action); $\mathbbm{1}_{A}=1$ holds when the event $A$ is \texttt{true} and otherwise $\mathbbm{1}_{A}=0$. 

Note that this safe RL formulation in~\eqref{eq.rcrl_problem} is different from the conventional one~\eqref{eq.srl_problem} in two ways: (1) \textit{the objective function} is separated into two parts where for states inside $\SPACE{S}_f^\pi$, it is possible to realize persistent safety so we can maximize return and guarantee safety simultaneously. However, for initial states outside $\SPACE{S}_f^\pi$, the desired constraint $h(s,a)\le0$ will be violated sooner or later so it is meaningless to optimize the return. We only try to minimize the worst violation; (2) \textit{the constraint} is imposed on every initial state inside the feasible set $\SPACE{S}_f^\pi$. Therefore, there are multiple (maybe infinite in continuous cases) constraints in the formulation rather than an expected one in~\eqref{eq.srl_problem}. Prior work~\cite{ma2022Joint, maFeasible} call this type constraints \textit{statewise} constraints and this formulation aims at safety at \textit{each time step} and \textit{each possible state}. The merits of reachability certificate in RL are discussed in~\cite{yu2022Reachabilitya} and we utilize it directly in this paper.

\section{Safe MBRL via Uncertainty-Aware Reachability Certificate}

Besides solving the constrained optimization problem~\eqref{eq.rcrl_problem}, we aim to reduce the training violation. To achieve this, we extend the reachability certificate to an uncertainty-aware formulation (i.e., a distributional certificate) as well as a shield policy considering the model uncertainty in Section~\ref{sec.drc} and~\ref{sec.shield}. Section~\ref{sec.practical} shows the practical algorithm. Specifically, we show how we learn and leverage the model in Section~\ref{sec.model_learning}. We implement the distributional reachability certificate (DRC) in a Gaussian form and impose constraints on its quantile in Section~\ref{sec.gaussian}. Section~\ref{sec.update_rules} gives the objective functions of each parameterized model while Algorithm~\ref{alg.DRPO} illustrates DRPO further.

\subsection{Distributional Reachability Certificate: Uncertainty-Aware Safety Critic}
\label{sec.drc}

\begin{figure}[!htb]
\centering
    \includegraphics[width=0.65\columnwidth]{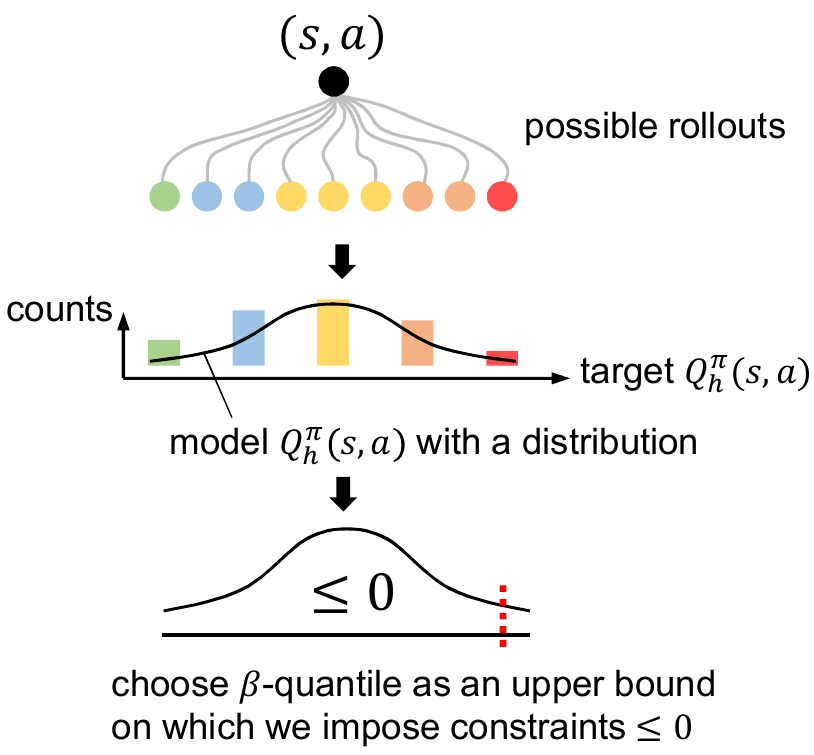}
    \caption{We illustrate the core idea of distributional reachability certificate. Due to the probabilistic model (either posterior sampling in~\cite{as2022Constraineda} or gaussian ensembles in~\cite{janner2019When}), the state trajectories vary a lot, leading to a distribution of $Q_h^\pi(s,a)$. Therefore, we approximate $Q_h^\pi(\cdot)$ with a distribution rather than only focusing its expectation. Our goal is to optimize the policy to shift the distribution and make  its $\beta$-quantile to be below zero. Then the possible trajectory is persistently safe with high probability no matter what the trajectory will be (i.e., a robust safety).}
    \label{fig.qh_distribution}
    \vspace{-1mm}
\end{figure}

One key issue that arises in MBRL is the model error~\cite{chua2018Deep}, which results in discrepancies between the predicted return under the learned model and the true dynamics. Although~\cite{chua2018Deep, janner2019When} proposed probabilistic model ensembles and clipped rollouts to cover the true dynamics within the support of the ensembles and mitigate the error, the stochasticity of the subsequent states $s'$ generated by the uncertain model $\hat{P}$ will still lead to a deviated or even wrong estimation about the cost value or reachability certificate in safe RL if not addressed properly. 
In the context of safety-critical problems, it is not enough to only care about the expected safety quantities of the predicted trajectories because an unsafe state misunderstood as safe will bring catastrophic failure to the system.

Furthermore, with the existence of a probabilistic model, the reachability certificate value $Q_h^\pi(\cdot)$ of a given state-action pair $(s,a)$ under the learned model is inevitably a random variable rather than a fixed quantity because the subsequent states and the subsequent state trajectories may be totally different during the model rollouts, as shown in Fig~\ref{fig.qh_distribution}.

Hence, we need a reachability certificate robust to the potential model error and aware of the uncertainty of the learned model. Inspired by recent progress in distributional RL (i.e., modeling the value functions in RL as distributions instead of focusing on its expected value)~\cite{bellemare2017Distributional, duan2021Distributional, ma2020DSAC}, we propose distributional reachability certificate. By imposing constraints on its $\beta$-quantile (\textit{a.k.a.} value-at-risk, VaR), we are able to get a safety quantity robust to the model error and uncertainty with a high confidence level. Furthermore, the robust reachability certificate is not conservative because the model accuracy is improved and the uncertainty is decreased as learning proceeds.

We consider to model the distribution of the reachability certificate instead of its expected value. We define $\mathcal{Z}(Q_h^\pi(s,a)|s,a): \SPACE{S}\times\SPACE{A} \mapsto \mathcal{P}(Q_h^\pi(s,a))$ as the mapping from $(s,a)$ to a distribution over the reachability certificate value and call it the distributional reachability certificate (DRC). Similar to~\cite{bellemare2017Distributional}, we define the \textit{distributional Bellman operator of reachability certificate} as
\begin{equation}
\begin{aligned}
\label{eq.bellman_drc}
\mathcal{T}^\pi Q_h(s,a) \overset{D}{\coloneqq} (1-&\gamma)h(s,a)\\
+&\gamma\max\left\{h(s,a), Q_h(s',a')\right\} \\
s'\sim\hat{P}(\cdot|s,a),~&a'\sim\pi(\cdot|s')
\end{aligned}
\end{equation}
where $A\overset{D}{=}B$ denotes that two random variables $A$ and $B$ shares the same probability density function. The uncertainty of $\mathcal{T}^\pi Q_h(s,a)\sim\mathcal{T}^\pi \mathcal{Z}(\cdot|s,a)$ consists of two parts under settings in this paper: (1) the probabilistic transition model $s'\sim\hat{P}(\cdot|s,a)$ and (2) the distributional next-state-action certificate $Q_h(s',a')$. Banach's fixed point theorem indicates that $\mathcal{T}^\pi$ has a unique fixed point $\mathcal{Z}(Q_h^\pi(s,a)|s,a)$ according to~\cite{bellemare2017Distributional}. To conclude, we can compute $\mathcal{Z}(Q_h^\pi(s,a)|s,a)$ by solving the optimization problem iteratively:
\begin{equation}
\label{eq.loss_q_h}
\mathcal{Z}_{\rm new} = \arg\min\limits_{\mathcal{Z}} \mathop{\mathbb{E}}\limits_{(s,a)\sim d_\pi} [D(\mathcal{T}^\pi\mathcal{Z}_{\rm old}(\cdot|s,a), \mathcal{Z}(\cdot|s,a))],
\end{equation}
where $D$ is some distance metric between two distributions. The Kullback–Leibler (KL) divergence $D_{\rm KL}$ is commonly adopted in distributional RL~\cite{duan2021Distributional, ma2020DSAC, bellemare2017Distributional}.

Suppose that the distribution $\mathcal{Z}(\cdot|s,a)$ covers the true reachability certificate value of the $(s,a)$ pair. In order to realize safety robust to the model uncertainty, we need to guarantee that most samples of $\mathcal{Z}(\cdot|s,a)$ adhere to the constraints (i.e., less than or equals to zero). Given a user-specified quantile fraction (\textit{a.k.a.} confidence level) $\beta\in(0,1]$ (ideally, $\beta$ is close to one), the $\beta$-quantile function (or VaR of $(1-\beta)$) of the DRC is defined as $F^{-1}_{Q_h^\pi}(\beta)\coloneqq\inf\{{q_h^\pi\in\mathbb{R}}: F_{Q_h^\pi}(q_h^\pi)\ge\beta\}$, where $F_{Q_h^\pi}(q_h^\pi)$ is the cumulative distribution function (CDF) of the distribution $\mathcal{Z}(\cdot|s,a)$ and $F^{-1}$ is its inverse; $Q_h^\pi$ is the random variable and $q_h^\pi$ is a specific value. The meaning of the $\beta$-quantile function of DRC lies in that once we guarantee  $F^{-1}_{Q_h^\pi(s,a)}(\beta) \le 0$, the trajectory starting from $(s,a)$ remains safe with at least the probability of $\beta$ as long as following $\pi$ because the true value $q_h^\pi(s,a)$ (the worst-case violation in the long term) is covered by $(-\infty, F^{-1}_{Q_h^\pi(s,a)}(\beta))$ with at least the probability of $\beta$. Thus, we get $\mathbb{P}\{Q_h^\pi(s,a)\le 0 \}\ge\beta$ if $F^{-1}_{Q_h^\pi(s,a)}(\beta) \le 0$. We denote $Z_\beta(s,a;\pi)\coloneqq F^{-1}_{Q_h^\pi(s,a)}(\beta)$ for simplicity from now on. Finally, all we need to do to realize a robust safety w.r.t. the model uncertainty is to replace the terms $Q_h^\pi(s,a)$ in~\eqref{eq.rcrl_problem} with $Z_\beta(s,a;\pi)$:
\begin{equation}
\begin{aligned}
\label{eq.drcrl_problem}
\max_{\pi\in\Pi} & \quad \mathbb{E}_{s\sim d_0(s)} \left[Q^{\pi}(s,a)\cdot\mathbbm{1}_{s\in \SPACE{S}_{\beta}^\pi} - Z_\beta(s,a;\pi) \cdot \mathbbm{1}_{s\notin \SPACE{S}_{\beta}^\pi}\right] \\
\mbox{s.t.} & \quad Z_\beta(s,a;\pi) \le 0, \forall s\in \SPACE{S}_0 \cap \SPACE{S}_{\beta}^\pi
\end{aligned},
\end{equation}
where $\SPACE{S}_{\beta}^\pi\coloneqq \left\{s:\min\limits_{a\in\SPACE{A}} Z_\beta(s,a;\pi)\le0\right\}$.

\subsection{Shield Policies and Safety Framework}
\label{sec.shield}
In practice, a intermediate solution policy of~\eqref{eq.drcrl_problem} may trade-off between safety and performance \textit{inappropriately} and tend to violate constraints for more rewards because of insufficient updates, which we want to mitigate in this work. Hence, we propose a shield policy \textit{minimizing} the $\beta$-quantile of long-term constraint violation (reachability certificate) and overwrite the action $a$ proposed by the original policy when the predicted $\beta$-quantile DRC of $(s,a)$ is beyond zero. We first give formal definitions of the shield policy.
\begin{definition}[$\beta$-shield policy]
An $\beta$-shield policy $\pi$ for $\mathcal{Z}(\cdot|s,a)$ minimizes the $\beta$-quantile of $Q_h^{\pi}$. The set of $\beta$-shield policies is
\begin{equation}
\begin{aligned}
\label{eq.optimal_shield_policy}
\Pi_{\beta} \coloneqq \{\pi\in\Pi: \sum_a \pi(a|s) Z_\beta(s,a;\pi) = \min_{a\in\SPACE{A}} Z_\beta(s,a;\pi) \}.
\end{aligned}
\end{equation}
\end{definition}
Analogous to the value iteration architecture in conventional RL~\cite{sutton2018Reinforcement}, a $\beta$-shield policy $\pi_h\in\Pi_{\beta}$ and its corresponding DRC can be computed by alternating between solving $\min_{\pi\in\Pi} \mathbb{E}[Z(s,a;\pi)]$ and solving $\min_{\mathcal{Z}}\mathbb{E}[D_{KL}(\mathcal{T}\mathcal{Z}, \mathcal{Z})]$, where $\mathcal{T}$ is the distributional Bellman \textit{optimality} operator of reachability certificate,
\begin{equation}
\nonumber
\mathcal{T} Q_h = \mathcal{T}^\pi Q_h, \mbox{for some }\pi\in\Pi_{\beta}.
\end{equation}

Once we get an intermediate $\pi_h$ and its DRC $\mathcal{Z}(\cdot|s,a)$, we will leverage it in exploration to reduce training time violations by invoking $\mathcal{Z}$ before taking action $a$. If $Z_\beta(s,a;\pi_h)>0$, following a safety-oriented policy $\pi_h$ will still violate constraints with a high probability after taking $a$ at $s$, so $a$ should be overwritten by a safer action, e.g. $\pi_h(s)$.

One possible issue arises from the overwriting action is the conservativeness because the action given by $\pi_h$ is safety-oriented and it will hinder the reward acquisition temporarily. We address this issue in two ways, both from the perspective of optimization problem and the one of execution, forming our safe RL framework illustrated in Fig~\ref{fig.framework}.

\textbf{From the optimization problem} \quad We replace the constraint $Z_\beta(s,a;\pi) \le 0$ in~\eqref{eq.drcrl_problem} with the constraints on the DRC of the \textit{shield policy}, leading to the final problem formulation of this work,
\begin{equation}
\begin{aligned}
\label{eq.formal_drcrl_problem}
\max_{\pi\in\Pi} & \quad \mathbb{E}_{s\sim d_0(s)} \left[Q^{\pi}(s,a)\cdot\mathbbm{1}_{s\in {\SPACE{S}_{\beta}}} - Z_\beta(s,a;{\pi_h}) \cdot \mathbbm{1}_{s\notin {\SPACE{S}_{\beta}}}\right] \\
\mbox{s.t.} & \quad Z_\beta(s,a;{\pi_h}) \le 0, \forall s\in \SPACE{S}_0 \cap {\SPACE{S}_{\beta}}, \mbox{for some }\pi_h\in\Pi_{\beta}
\end{aligned},
\end{equation}
where $\SPACE{S}_{\beta} \coloneqq \left\{s:\min_{a\in\SPACE{A}} Z_\beta(s,a;\pi_h)\le0\right\}$ and the main difference is that all constraint parts are related to the shield policy. The merit of taking $Z_\beta(s,a;\pi_h)$ as the constraint lies in that $Z_\beta(s,a;\pi_h) \le Z_\beta(s,a;\pi), \forall \pi\in\Pi$. Therefore, it is easy to obtain $\SPACE{S}_{\beta}^\pi \subseteq \SPACE{S}_{\beta}$, which means the agent is allowed to work in a \textit{larger} subspace of $\SPACE{S}$. This is because the long-term violation of $(s,a)$ will be decreased by taking $\pi_h$, a safety-only policy, as the future policy. This approach is also similar with the least-restrictive method~\cite{kang2022lyapunov} due to a larger workspace.

\textbf{From the policy execution} \quad As shown in Fig~\subref*{fig.overview_shield}, there may be a margin between the safe action $a_h$ and the action $a$ given by $\pi$. Therefore, applying $a_h$ directly will lose the optimality. We propose a line search method between $a_h$ and $a$ in the action space, i.e., finding the nearest coarse action $a_t\coloneqq k a + (1-k) a_h$ whose DRC is below zero, where the optimal $k^*$ is given by
\begin{equation}
\begin{aligned}
\label{eq.line_search}
k^* = {\min} \{k\in &\{0,\frac{1}{N}, \dots, \frac{N-1}{N}, 1\}: \\
&Z_\beta(s, k a + (1-k) a_h;\pi_h)\le0\}.
\end{aligned}
\end{equation}

We find $N=10$ is sufficient for tasks in this work. This simple approach resembles the projection method in~\cite{yang2020ProjectionBased} but the line search trick is simpler and we find it effective during execution. Note that we adopt the line search method only when evaluating the intermediate policy, and we still overwrite $a$ with $a_h$ if unsafe (i.e., $Z_\beta(s,a;\pi_h;\psi)>0$) during training. The whole process of DRPO is illustrated in Fig~\ref{fig.framework}\subref{fig.overview_shield}, including an RL problem constraining the distributional reachability certificate, a $\beta$-shield policy and the line search method, forming a complete safe RL framework.

\begin{algorithm*}[!htp]
\caption{Distributional Reachability Policy Optimization}
\label{alg.DRPO}
\begin{algorithmic}[1]
\Require Rollout length $H$, episode length $T_{\mathrm{ep}}$, gradient descent steps $N_{\rm grad}$
\State Initialize main policy $\pi_{\theta}$, critic $Q^\pi_{\omega}$, DRC $\mathcal{Z}_\psi$, shield policy $\pi_{h_\nu}$, multiplier $\lambda_\xi$; and empty buffer $\mathcal{D}_{\mathrm{real}}$ and $\mathcal{D}_{\mathrm{virt}}$; the ensemble of learned models $\{\hat{P}_{\phi_i}\}_{i=1}^B$
\For{episode 1, 2, \dots}
    \For{$T_\mathrm{ep}$ times}
        \State{Interact with $\pi_\theta$ and $\pi_{h_\nu}$ by checking the $\beta$-quantile $Z_\beta(s,a;\pi_h;\psi)$ $\overset{\text{?}}{\le}$ $0$}; add samples to $\mathcal{D}_{\mathrm{real}}$ \Comment{Interaction}
        \State{Fit models $\{\hat{P}_{\phi_i}\}_{i=1}^B$ by performing mini-batch GD on~\eqref{eq.loss_model} with $\mathcal{D}_{\mathrm{real}}$} \Comment{Model Learning}
        \State{Sample $s\sim\mathcal{D}_{\mathrm{real}}$}
        \State{Rollout from $s$ for $H$ steps with $\pi_\theta$ and $\{\hat{P}_{\phi_i}\}_{i=1}^B$; add the samples to $\mathcal{D}_\mathrm{virt}$} \Comment{Model rollout}
        
        \For{$N_{\rm grad}$ times}
            \State{Sample mini-batch from $\mathcal{D}_\mathrm{virt}$ for successive updates}
            \State{Update $Q_\omega$ by mini-batch GD on~\eqref{eq.loss_q}; Update $\mathcal{Z}_\psi$ by mini-batch GD on~\eqref{eq.kl_loss_q_h}} \Comment{Critics learning}
            \State{Update $\pi_\theta$ by mini-batch GD on~\eqref{eq.loss_pi_and_lambda}; Update $\pi_{h_\nu}$ by mini-batch GD on~\eqref{eq.loss_pi_h}}  \Comment{Actors learning}
            \State{Update $\lambda_\xi$ by mini-batch GD on~\eqref{eq.loss_pi_and_lambda}} \Comment{Multiplier learning}
        \EndFor
    \EndFor
\EndFor
\end{algorithmic}
\end{algorithm*}

\subsection{Practical Implementation}
\label{sec.practical}

\subsubsection{Model Learning and Usage} 
\label{sec.model_learning}

Same as prior MBRL work~\cite{chua2018Deep, janner2019When, thomas2021Safe}, we adopt an ensemble of diagonal Gaussian dynamics model parameterized by $\phi$ as the world model approximators, denoted as $\{ \hat{P}_{\phi_i} \}_{i=1}^B$, where $\hat{P}_{\phi_i} = \mathcal{N}(\mu_{\phi_i}(s,a), \sigma^2_{\phi_i}(s,a))$. The models are updated via maximum likelihood on all historical transitions $(s,a,r,s',h)$ from the replay buffer $\mathcal{D}_{\rm real}$:
\begin{equation}
\label{eq.loss_model}
\mathcal{J}_{\hat{P}}(\phi_i) = -\mathop{\mathbb{E}}\limits_{(s,a,r,s',h)\sim\mathcal{D}_{\mathrm{real}}} \log \hat{P}_{\phi_i}(s',r\mid s,a).
\end{equation}
Note that the constraint function can either be known \textit{a priori} or be learned similarly as the reward function. We test both cases in Section~\ref{sec.exp}. Each model in the ensemble is initialized randomly and updated with different mini-batches selected from the buffer $\mathcal{D}_{\rm real}$, resulting in totally different models and rollouts given the same starting point $(s,a)$.

The models are utilized to generate virtual transitions for learning the critics and policies. At each time step, we sample a mini-batch from the buffer as the starting point and perform one-step prediction by choosing one model $\hat{P}_i$ randomly from the ensemble. After repeating the rollout for $H$ times, where $H$ is a hyper-parameter, we add the transitions to a virtual buffer $\mathcal{D}_{\rm virt}$ for updating the critics and policies. This type of \textit{truncated} rollout method leads to a smaller error in terms of the value function~\cite{janner2019When}. But its probabilistic nature still brings distributional predictions that need to be addressed, especially under safety-critical circumstances.

\subsubsection{Gaussian Approximated DRC Learning}
\label{sec.gaussian}
Despite the shape of the true reachability certificate, we can approximate it with parameterized and tractable distribution, e.g., a diagonal Gaussian. As long as the mean of the approximated Gaussian is close to the true certificate value, the $\beta$-quantile of the Gaussian will cover the true value with high-probability, leading to a safety guarantee. This can be achieved by model learning, which brings a decreasing discrepancy between the certificate of the learned model and the one under the true dynamics. However, a Gaussian may still be a coarse approximation of a complicated distribution as mentioned in~\cite{yang2022Safetyconstrained} and interested readers are referred to~\cite{yang2021WCSAC, yang2022Safetyconstrained} for advanced approaches.

We denote the parameterized Gaussian approximation of DRC as $\mathcal{Z}_\psi(\cdot|s,a)\coloneqq\mathcal{N}(\mu_\psi(s,a),\sigma_\psi^2(s,a))$ with parameter $\psi$. For example, the mean and the variance can be approximated by two neural networks parameterized by $\psi$. Then the model-based KL-divergence version of~\eqref{eq.loss_q_h} is
\begin{equation}
\begin{aligned}
\label{eq.kl_loss_q_h}
&\mathcal{J}_\mathcal{Z}(\psi) = \mathop{\mathbb{E}}\limits_{\tiny (s,a)\sim\mathcal{D}_{\rm virt}} \left[ D_{\rm KL}(\mathcal{T}\mathcal{Z}_{\psi'}(\cdot|s,a), \mathcal{Z}_\psi(\cdot|s,a)) \right] \\
&= c-\mathop{\mathbb{E}}_{\tiny (s,a)\sim\mathcal{D}_{\rm virt}, a'\sim \pi_h \atop
      Q_h(s,a)\sim \mathcal{Z}_{\psi'}(\cdot|s,a)
    }
  \left[ \log \mathcal{P}\left( \mathcal{T}Q_h(s,a) | \mathcal{Z}_\psi(\cdot|s,a)) \right) \right],
\end{aligned}
\end{equation}
where $c$ is a term independent of $\psi$ and detailed derivation can be found in~\cite{duan2021Distributional}. By minimizing~\eqref{eq.kl_loss_q_h}, we obtain a Gaussian approximation of $\mathcal{Z}(\cdot|s,a)$ and its $\beta$-quantile is
\begin{equation}
\label{eq.gaussian_quantile}
Z_\beta(s,a) = \mu_\psi(s,a) + \Phi^{-1}(\beta) \sigma_\psi(s,a),
\end{equation}
where $\Phi^{-1}(\beta), \beta\in(0,1)$ is the inverse function of CDF of standard normal distribution. Common $(\beta, \Phi^{-1}(\beta))$ pairs include $(0.841, 1), (0.977, 2)$ and $(0.999, 3)$ and we compare their corresponding efficacy in the ablation study.

\subsubsection{Model-based Lagrangian Optimization}
\label{sec.update_rules}
We leverage the widely used Lagrange multiplier method to solve the constrained optimization problem~\eqref{eq.formal_drcrl_problem}. The problem is transformed into solving the saddle point of Lagrangian $\mathcal{L}(\pi, \lambda)$. Similar to ~\cite{yu2022Reachabilitya}~\cite{maFeasible}, the statewise constraints are imposed on each state in $\SPACE{S}_0 \cap \SPACE{S}_{\beta}$, so the Lagrangian is formulated as
\begin{equation}
\begin{alignedat}{2}
\label{eq.Lagrangian_drcrl_problem}
\mathcal{L}(\pi, \lambda) &= \mathbb{E}_{s\sim d_0(s)} [ -Q^{\pi}(s,a)\cdot\mathbbm{1}_{s\in \SPACE{S}_{\beta}} \\
&\quad\quad\quad\quad\quad+Z_\beta(s,a;\pi_h) \cdot \mathbbm{1}_{s\notin \SPACE{S}_{\beta}}] \\
&+\int_{\SPACE{S}_0 \cap \SPACE{S}_{\beta}}\lambda(s) Z_\beta(s,a;\pi_h)\mathrm{d}s,
\end{alignedat}
\end{equation}
where $\lambda:\SPACE{S}\mapsto[0,\lambda_{\mathrm{max}}]$ is the statewise multiplier function. We do not optimize (\ref{eq.Lagrangian_drcrl_problem}) directly due to the intractability of $\SPACE{S}_\beta$ in advance but a surrogate version that is easier to handle. The surrogate shares the same optimal solution with~\eqref{eq.Lagrangian_drcrl_problem} according to~\cite{yu2022Reachabilitya}:
\begin{equation}
\begin{aligned}
\label{eq.surrogate_Lagrangian_drcrl_problem}
\hat{\mathcal{L}}(\pi, \lambda) =  &\mathop{\mathbb{E}}\limits_{s\sim d_0(s) \atop a\sim \pi(\cdot|s) } \left[-Q^{\pi}(s,a) + \lambda(s)Z_\beta(s,a;\pi_h)\right].
\end{aligned}
\end{equation}
Therefore, (\ref{eq.formal_drcrl_problem}) can be solved by finding the saddle point of the surrogate Lagrangian (\ref{eq.surrogate_Lagrangian_drcrl_problem}) by gradient descent (GD):
\begin{equation}
\begin{aligned}
\label{eq.min_max_Lagrangian}
\min_{\pi}\max_{\lambda}\hat{\mathcal{L}}(\pi, \lambda).
\end{aligned}
\end{equation}
For the overall structure of DRPO, we adopt the popular actor-critic framework with deep neural networks as the approximators. In particular, we have two actor-critic structures, one for solving the $\beta$-shield policy $\pi_h(\cdot|s;\nu)$ and its corresponding DRC $\mathcal{Z}(\cdot|s,a;\pi_h;\psi)$, and the other for solving the main policy $\pi(\cdot|s;\theta)$, the Q-value function $Q(s,a;\omega)$ and the multiplier $\lambda(s;\xi)$. The DRC is updated by taking SGD on (\ref{eq.kl_loss_q_h}) while the $\beta$-shield policy $\pi_h(\cdot|s;\nu)$ tries to minimize the $\beta$-quantile of DRC
\begin{equation}
\begin{aligned}
\label{eq.loss_pi_h}
\mathcal{J}_{\pi_h}(\nu) = \mathop{\mathbb{E}}\limits_{s\sim \mathcal{D}_{\rm virt}\atop a\sim \pi_{h}(\cdot|s;\nu)} \left[Z_\beta(s,a;\pi_h;\psi)\right].
\end{aligned}
\end{equation}
The Q-value function is updated by minimizing the mean squared error between the current and target value:
\begin{equation}
\label{eq.loss_q}
\mathcal{J}_{Q}(\omega) = \mathop{\mathbb{E}}\limits_{(s,a,r,s')\sim \mathcal{D}_{\rm virt} \atop a'\sim \pi_\theta(\cdot|s')} \left[ Q(s,a;\omega) - (r +\gamma Q(s',a';\omega)) \right]^2.
\end{equation}
As mentioned earlier, the main policy $\pi(\cdot|s;\theta)$ is updated by descending
the surrogate Lagrangian while the multiplier $\lambda_\xi$ tries to ascend it:
\begin{equation}
\begin{aligned}
\label{eq.loss_pi_and_lambda}
\mathcal{J}_\pi(\theta) &= -\mathcal{J}_\lambda(\xi) \\
&= \mathop{\mathbb{E}}\limits_{s\sim \mathcal{D}_{\rm virt}\atop a\sim \pi(\cdot|s;\theta)} \left[-Q(s,a;\omega) + \lambda(s;\xi)Z_\beta(s,a;\pi_h;\psi)\right].
\end{aligned}
\end{equation}

We adopt soft-actor-critic (SAC)~\cite{haarnoja2018Soft}, a popular off-policy RL algorithm, as the policy optimizer. Thus, a regularization term related to the policy entropy is added to $\mathcal{J}_\pi(\theta)$. A complete version of DRPO is summarized in Algorithm~\ref{alg.DRPO}. Note that sometimes we take the parameters ($\theta, \nu, \omega, \psi, \phi, \xi$) as the subscript of the notation for simplicity.

\section{Experiments}
\label{sec.exp}
We evaluate our methods DRPO on classical robotics tasks such as constrained stabilization, tracking and sensorimotor navigation. Our goal is to validate: (1) whether DRPO is able to learn a safe policy with non-trivial performance; (2) whether the proposed DRC and the corresponding shield policy help to reduce training time violations.
\subsection{Environments}

\begin{figure}[hbt]
\vspace{-5mm}
\centering
    \subfloat[{Cartpole-Move}]{
        \includegraphics[width=0.3\columnwidth]{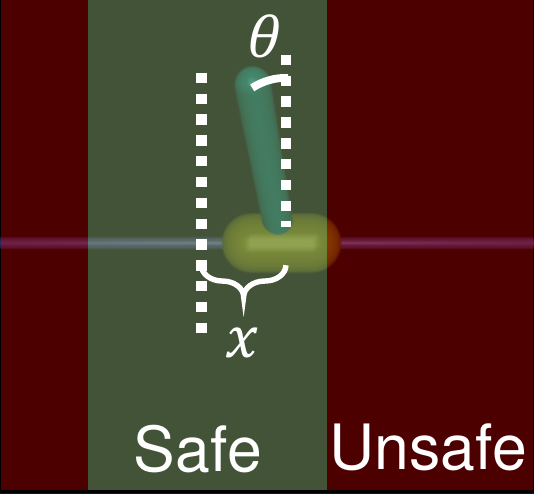}
        \label{fig.env_cartpole}
    }
    \hfill
    \subfloat[{Quadrotor}]{
	    \includegraphics[width=0.3\columnwidth]{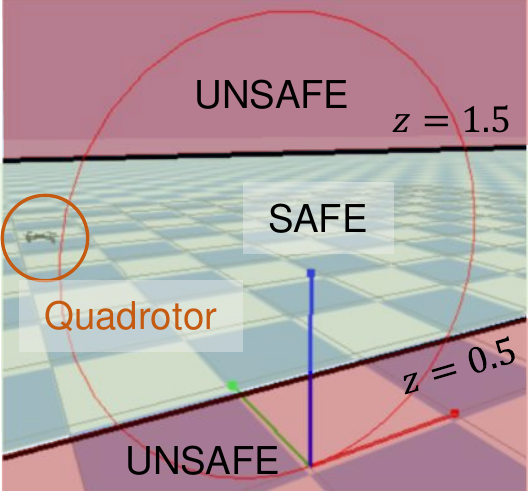}
        \label{fig.env_quadrotor}
    }
    \hfill
    \subfloat[{Safety-Gym}]{
	    \includegraphics[width=0.28\columnwidth]{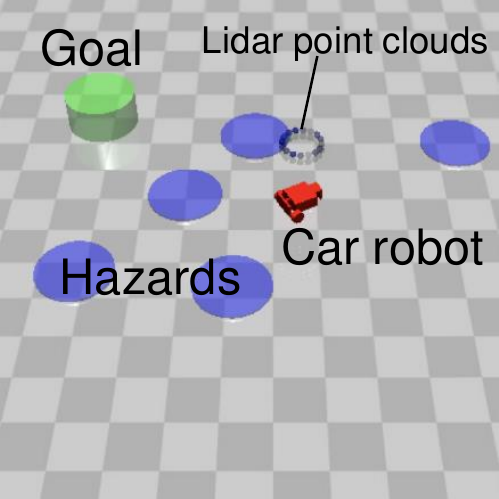}
        \label{fig.env_sg}
    }
    \caption{Snapshots of the three tasks, including agents and constraints.}
    \label{fig.envs}
    \vspace{-1mm}
\end{figure}
We test all algorithms on four environments in three benchmarks. Note that each episode will be terminated if the constraint is violated during training unless further explained.

\textbf{Cartpole-Move} is a task based on \textbf{cartpole} in OpenAI-Gym~\cite{brockman2016gym} and MuJoCo~\cite{todorov2012MuJoCo}, borrowed from~\cite{luo2021Learning}. The goal is to move the cart (yellow part in Fig~\ref{fig.env_cartpole}) to control the pole (blue part in Fig~\ref{fig.env_cartpole}). Starting from $(\theta, x)=(0,0)$, the system is constrained in the space $\{(\theta, x): |x|\le0.9, |\theta|\le0.2\}$, i.e., $h(s) = [x-0.9, -0.9-x, \theta-0.2,-0.2-\theta]^T$ (here we temporarily use $\theta$ to denote the deviation angle of the pole). The reward function is $r(s,a)=x^2$. Therefore, there is a conflict between obtaining rewards and staying safe. The cart should move slowly towards one direction and then stay still around $x<0.9$. The length of an episode $T_{\rm{ep}}$ is 1000.

\begin{figure*}[!htp]
\centering
\captionsetup[subfloat]{labelsep=none,format=plain,labelformat=empty}
\subfloat[]{\label{fig:cartpole_ret}\includegraphics[width=0.24\textwidth]{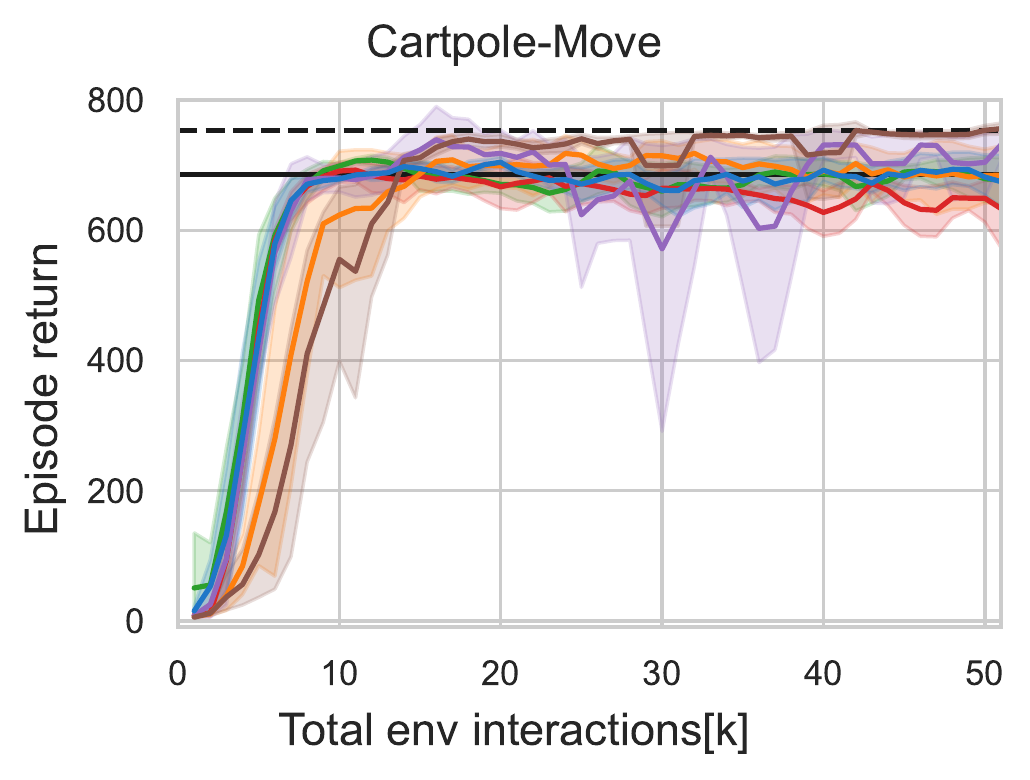}}\hfill
\subfloat[]{\label{fig:quadrotor_ret}\includegraphics[width=0.24\textwidth]{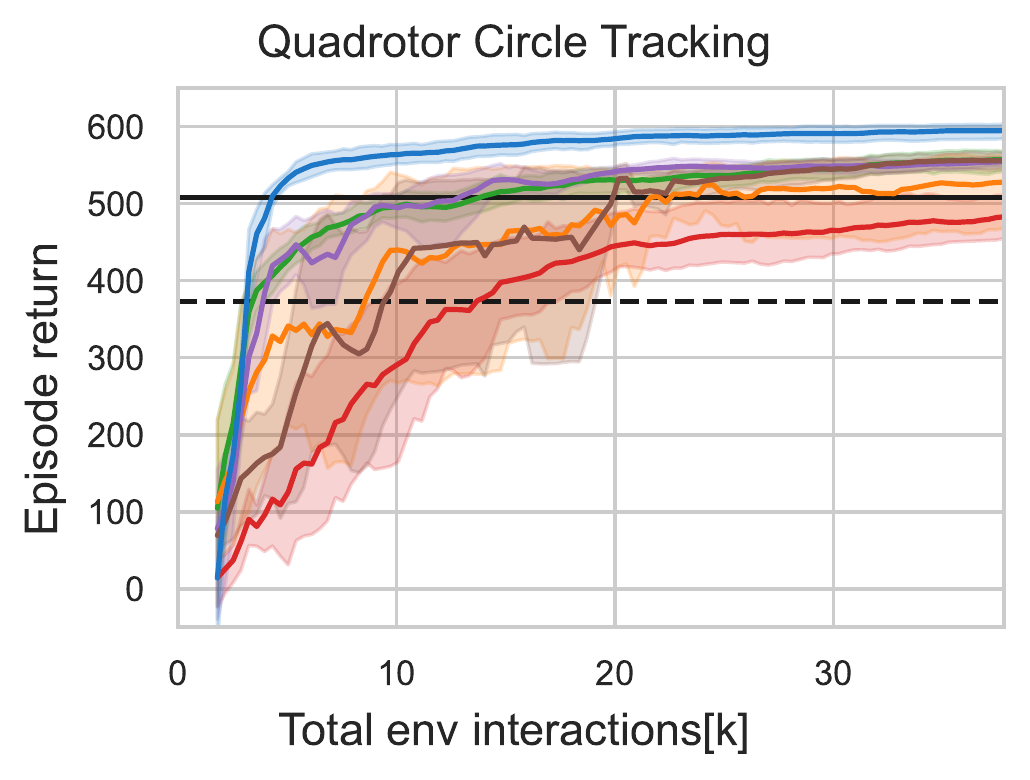}}\hfill
\subfloat[]{\label{fig:sgcar_ret}\includegraphics[width=0.24\textwidth]{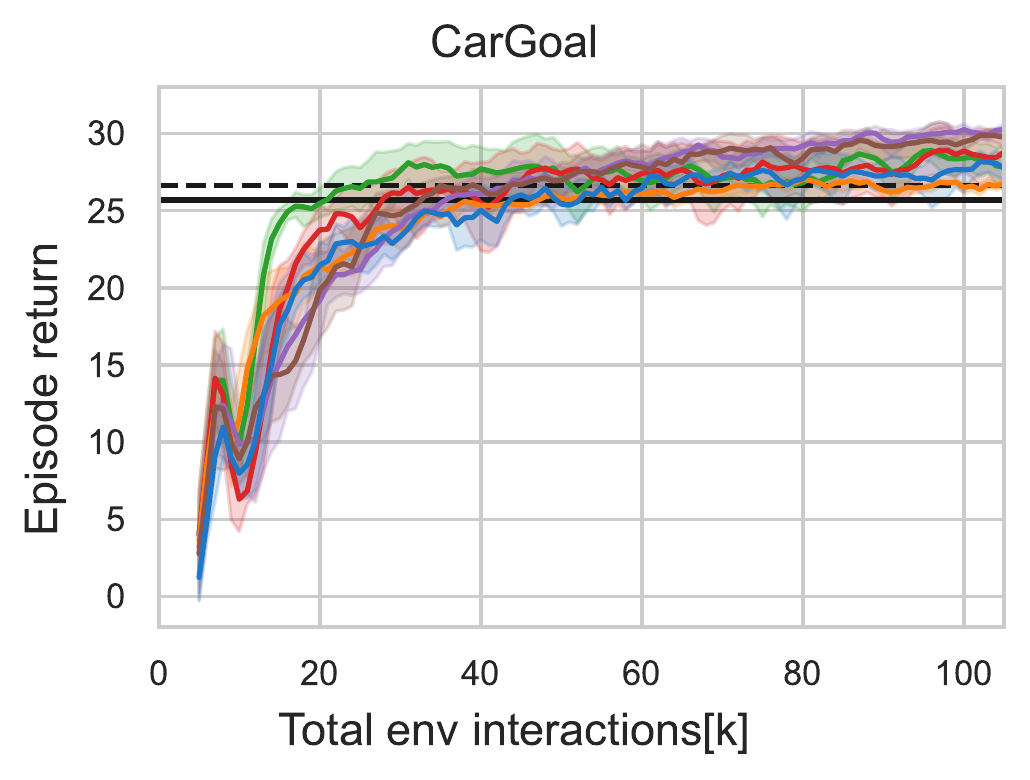}}\hfill
\subfloat[]{\label{fig:sgpoint_ret}\includegraphics[width=0.24\textwidth]{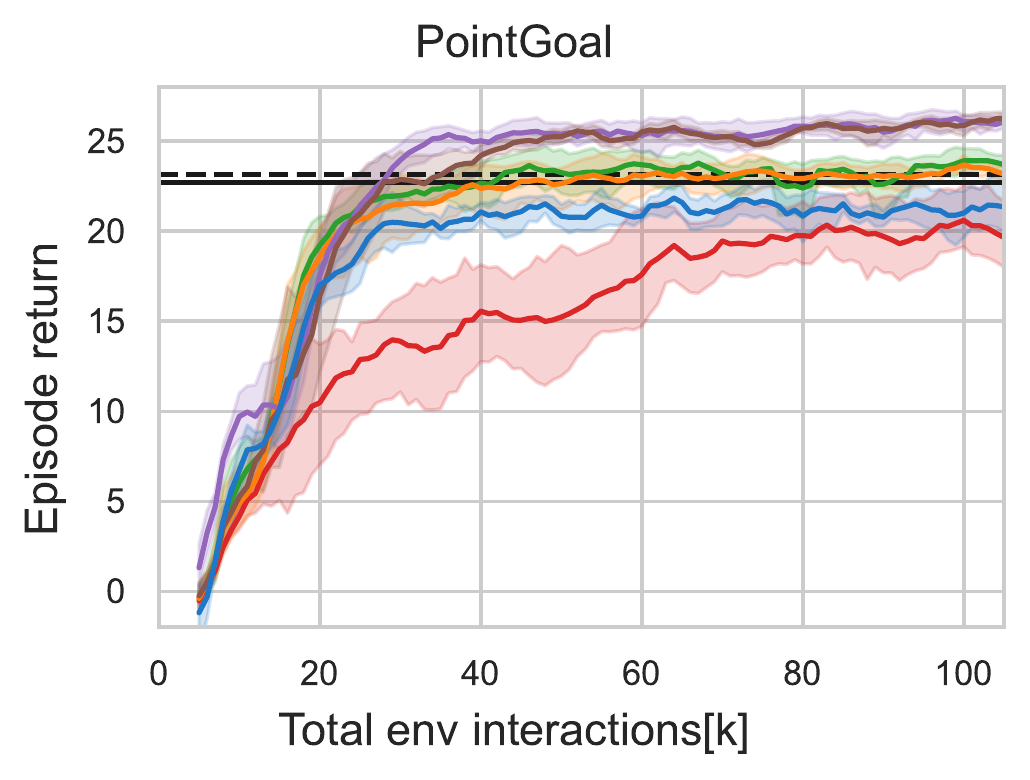}}
\vspace{-7mm}
\\
\subfloat[]{\label{fig:cartpole_vio_eval}\includegraphics[width=0.24\textwidth]{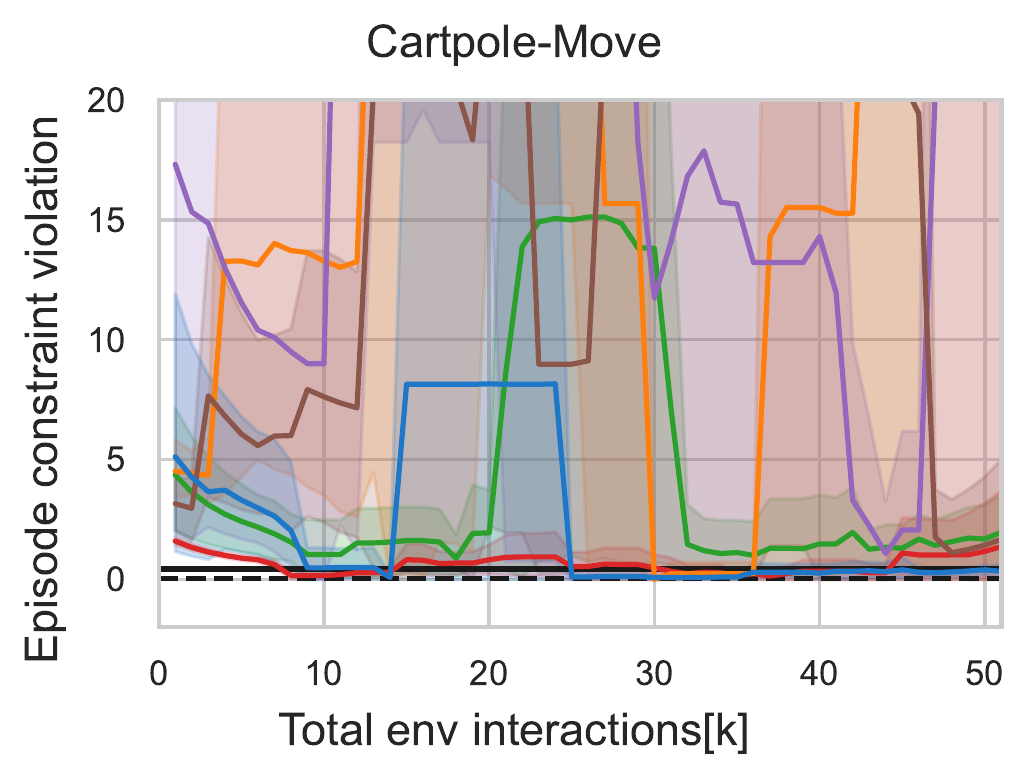}} \hfill
\subfloat[]{\label{fig:quadrotor_vio_eval}\includegraphics[width=0.24\textwidth]{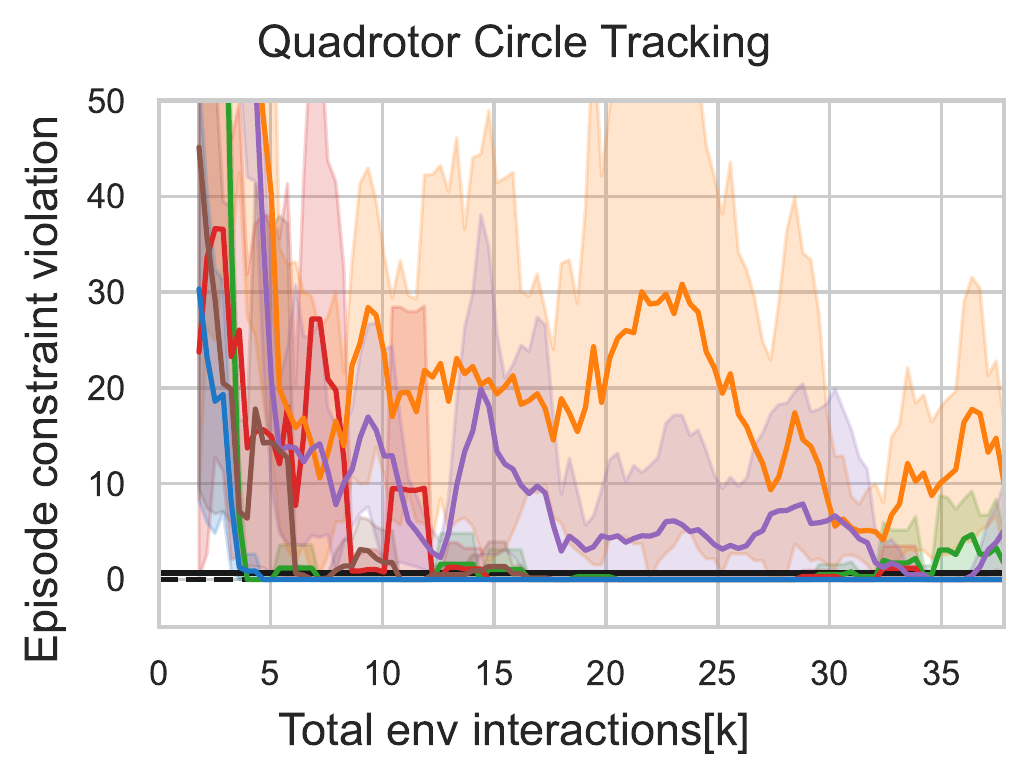}} \hfill
\subfloat[]{\label{fig:sgcar_vio_eval}\includegraphics[width=0.24\textwidth]{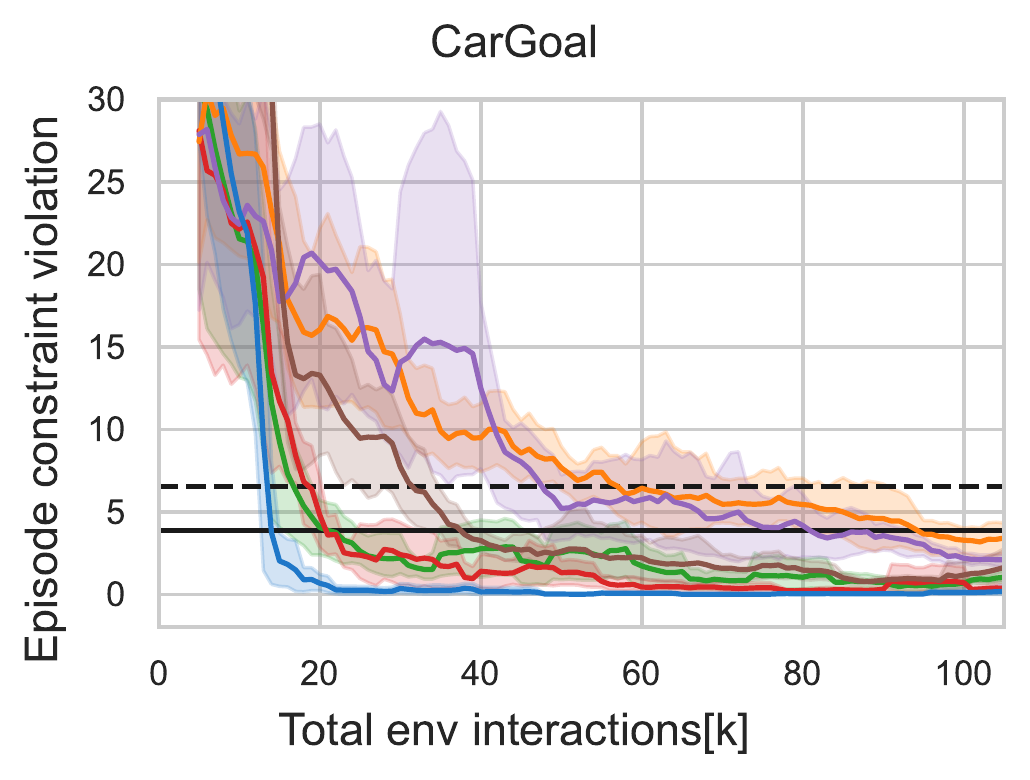}}\hfill
\subfloat[]{\label{fig:sgpoint_vio_eval}\includegraphics[width=0.24\textwidth]{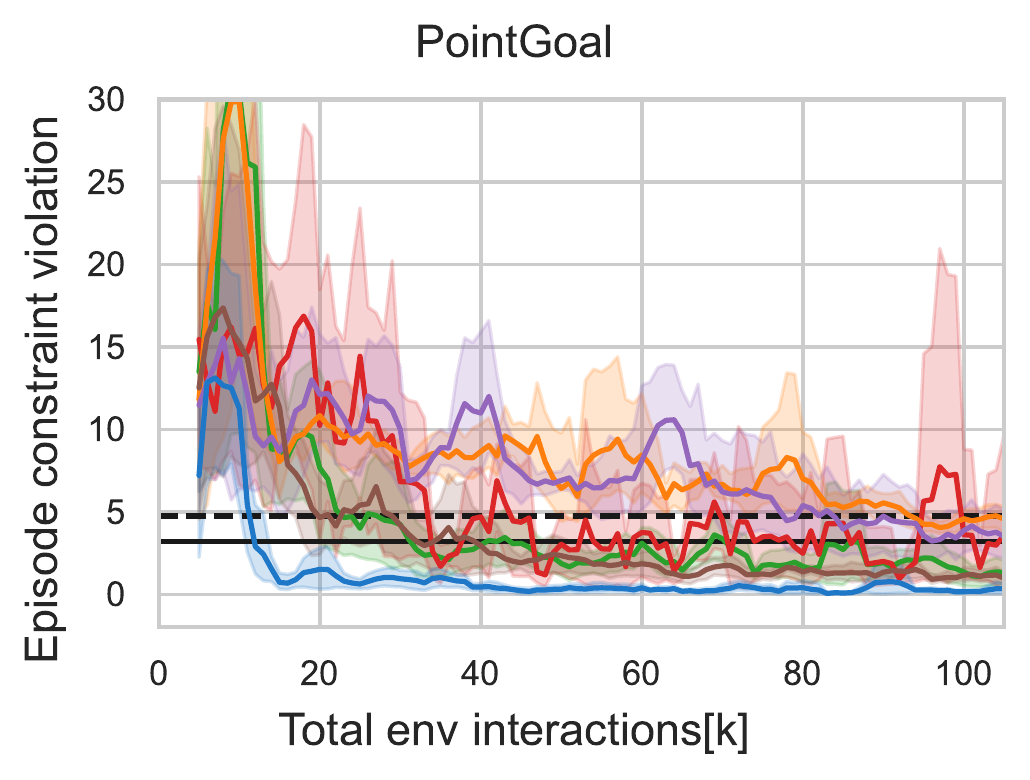}}\hfill
\vspace{-7mm}
\\
\subfloat[]{\label{fig:cartpole_vio_train}\includegraphics[width=0.24\textwidth]{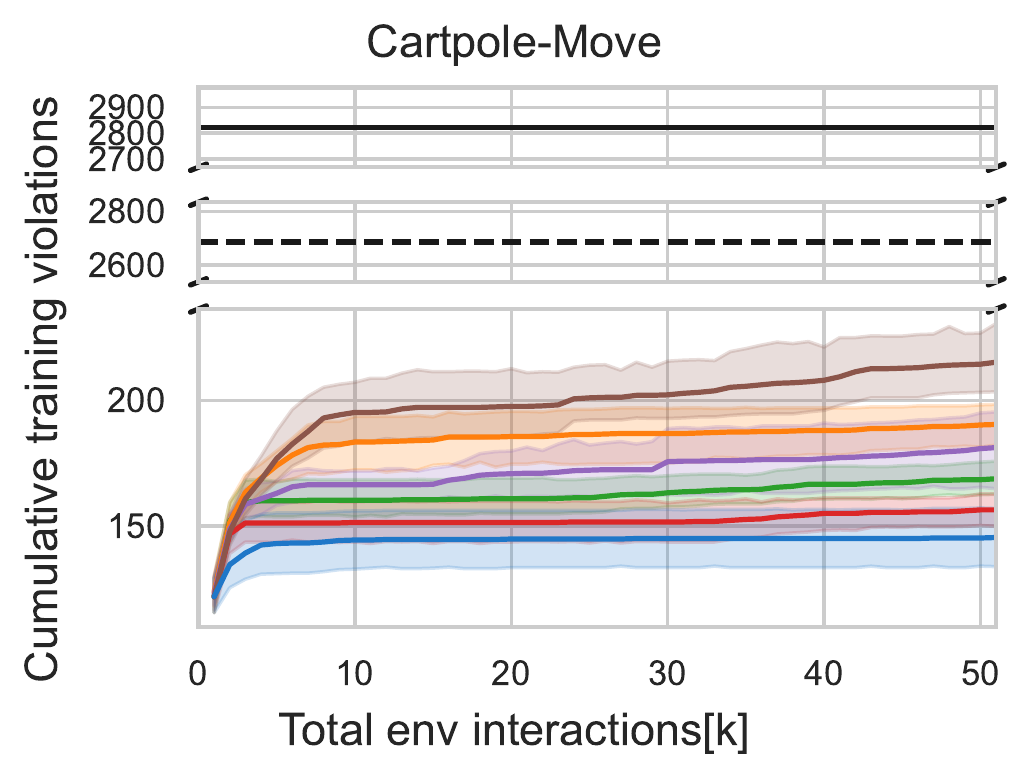}} \hfill
\subfloat[]{\label{fig:quadrotor_vio_train}\includegraphics[width=0.24\textwidth]{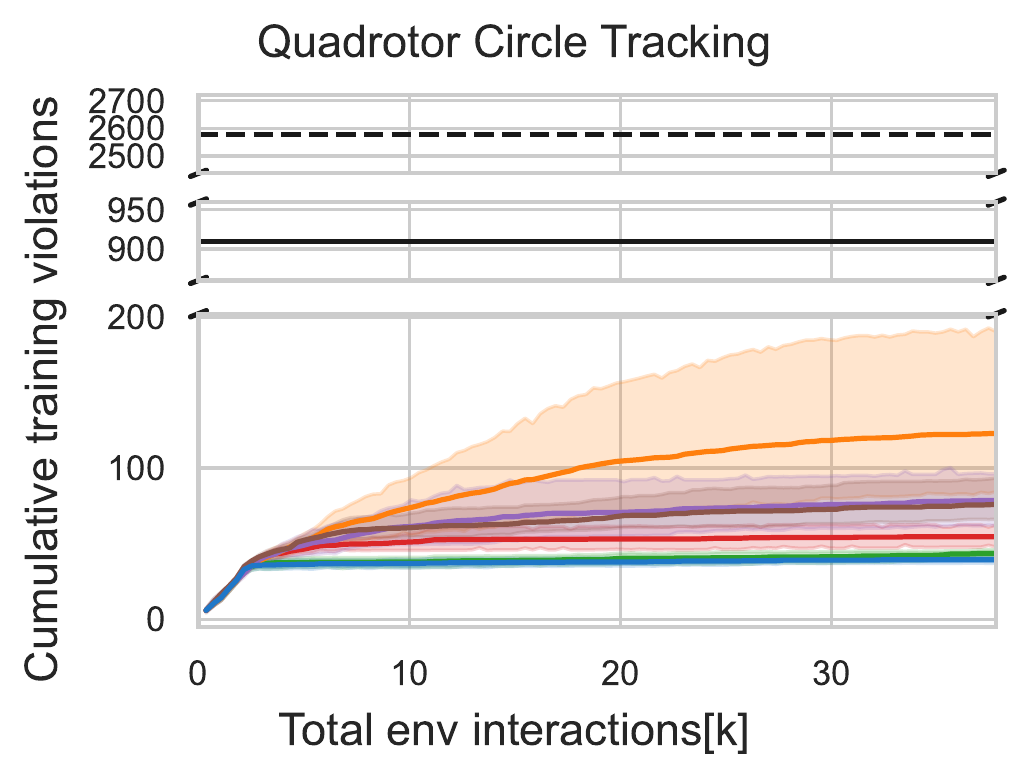}} \hfill
\subfloat[]{\label{fig:sgcar_vio_train}\includegraphics[width=0.24\textwidth]{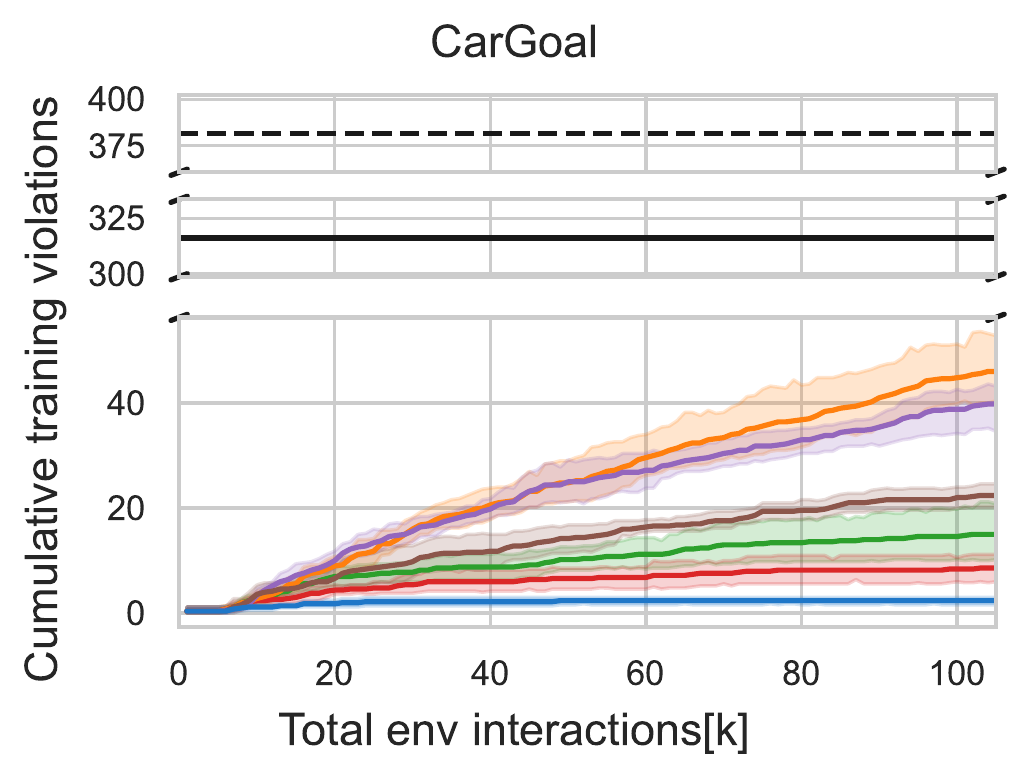}}\hfill
\subfloat[]{\label{fig:sgpoint_vio_train}\includegraphics[width=0.24\textwidth]{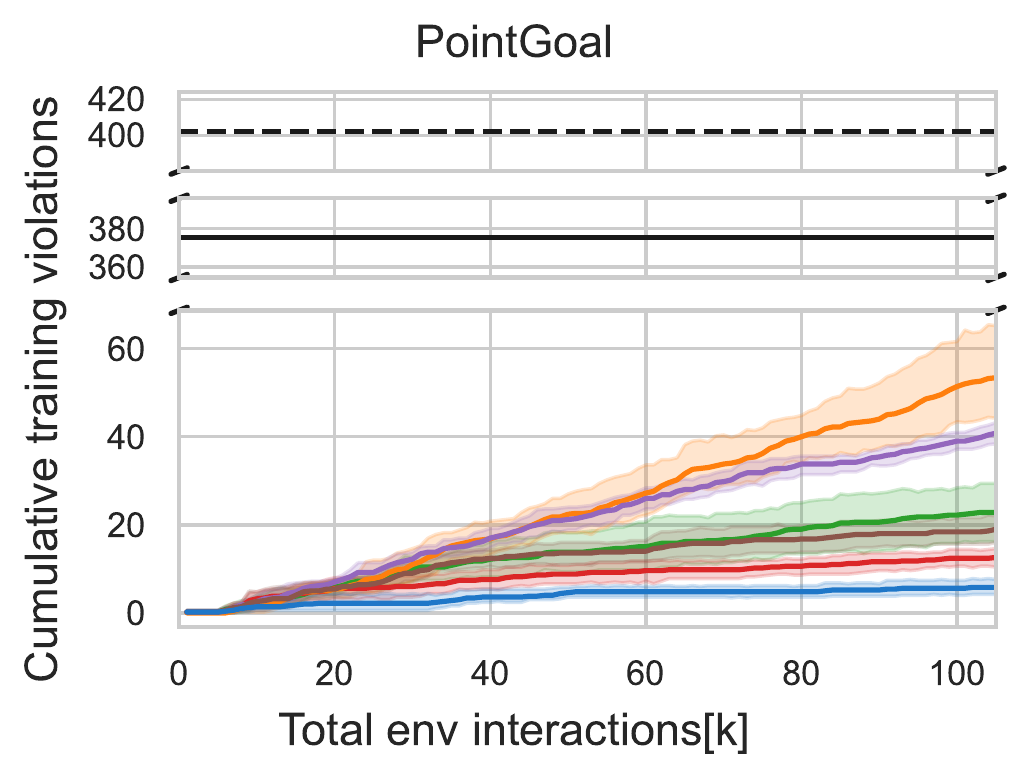}}
\vspace{-7mm}
\\
\subfloat[]{\includegraphics[width=0.5\textwidth]{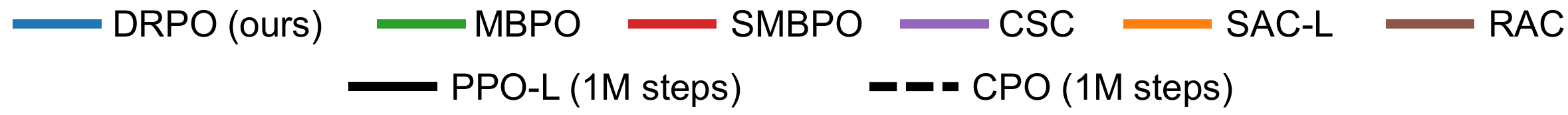}}
\vspace{-3mm}
\caption{Training curves on the four benchmarks. The solid lines correspond to the mean and the shaded regions correspond to 95\% confidence interval over 5 seeds. Each row corresponds one metric, from top to bottom, episode return, episode violation counts and cumulative training violations, respectively. Each column corresponds to the three metrics of one task.}
\label{fig.comparison_curves}
\vspace{-3mm}
\end{figure*}

\textbf{Quadrotor} is a 2D quadrotor tracking task in safe-control-gym~\cite{yuan2021safecontrolgym}. The policy needs to output normalized torque to control the quadrotor to track the circle trajectory (red line in Fig~\ref{fig.env_quadrotor}) as accurately as possible. However, the quadrotor is constrained w.r.t. its height $z$ where $h(s) = [z-1.5, 0.5-z]^T$ (black straight line). Therefore, the constraint contradicts the reward as well. The length of an episode $T_{\rm{ep}}$ is 360.

\textbf{CarGoal} and \textbf{PointGoal} come from {Safety-Gym}~\cite{ray2019Benchmarking}, a constrained robot navigation benchmark. The robot (red in Fig~\ref{fig.env_sg}) should reach the goal region while avoiding entering the blue hazards. The dynamics of car is more complex than ones of point. The layout of the environment is initialized randomly before each episode begins or after the robot reaches the goal. There are five hazards in the layout so the constraint is $h(s)=\max_{1\le i\le5}\{-\mathrm{dist}(\mathrm{robot}, \mathrm{hazard}_i)\}$, where $\rm dist$ is the signed distance between two objects. The reward is the difference of the distance between the robot and the goal of two time steps. These two tasks are more complicated than the other two because besides sensor readings and goal position, the robot have only access to \textit{lidar point clouds} of the hazards instead of their exact position. Therefore, the constraint function and the state of the robot should be learned. The length of an episode $T_{\rm{ep}}$ is 1000.

\subsection{Baselines}
The baselines for comparison include both model-free and model-based (safe) RL algorithms. 
We choose five model-free algorithms: \textbf{Constrained Policy Optimization} (CPO)~\cite{achiam2017Constrained} approximates the objective function and the constraint using trust region methods, and analytically solves for the policy parameters. \textbf{PPO-Lagrangian} (PPO-L)~\cite{ray2019Benchmarking} combines PPO with the Lagrangian method, which takes a weighted sum of the value and cost value function as the policy objective, and updates the multiplier using dual ascent. \textbf{SAC-Lagrangian} (SAC-L, \textit{a.k.a.} Reward Constrained Policy Optimization, RCPO) combines SAC with the Lagrangian method similarly with PPO-L. \textbf{Conservative Safety Critic} (CSC)~\cite{bharadhwaj2021Conservative} learns a safety critic that overestimates the probability of failure and uses it to constrain the policy improvement through primal-dual gradient descent. \textbf{Reachable Actor Critic} (RAC)~\cite{yu2022Reachabilitya} leverages the self-consistency condition to learn the expected reachability certificate, which characterizes the largest feasible set, and uses it to constrain the policy optimization.
We also choose two model-based algorithms. \textbf{Model-Based Policy Optimization} (MBPO)~\cite{janner2019When} uses short model-generated rollouts branched from real data to update the policy, and does not consider safety constraints. \textbf{Safe MBPO} (SMBPO)~\cite{thomas2021Safe} builds on MBPO and heavily penalizes unsafe trajectories to avoid safety violations. Our DRPO relies on MBPO as well but leverages DRC and the shield policy to reduce violations. We set $\Phi^{-1}(\beta)$ to $2$ in {Cartpole-Move} and {Quadrotor} and to $1$ in {Safety-Gym} tasks.

There are also safe RL approaches we do not compare with since they assume additional knowledge or other reasons. Recovery RL~\cite{thananjeyan2021Recovery} utilizes an offline dateset to pre-train a recovery policy and a recovery zone to guard the training process. Zhao et al.~\cite{zhao2021Modelfree} assume access to an accurate black-box model during planning, which is difficult to get in real-world. Luo et al.~\cite{luo2021Learning} and Berkenkamp et al.~\cite{berkenkamp2016Safe, berkenkamp2019Safe} start from an initially safe policy to guarantee safe learning, which is hard to obtain in complex tasks. Distributional functions in safe RL have been investigated recently in ~\cite{yang2021WCSAC, yang2022Safetyconstrained, kim2022TRC} but they focus on the stochasticity of model-free settings, which may cause more violations during interaction. In constrast, we propose DRC to address the model-uncertainty in model-based algorithms which leads to fewer violations in nature so we do not list their results in this work. Furthermore, DRC and the learned model in this work can be adopted in broader approaches such as model-based planning (MPC).

\subsection{Results}
We evaluate all algorithms with three metrics: (1) \textbf{episode return} (ER) $\sum_t^{T_{\rm ep}} r_t$ indicates the performance of different intermediate agents; (2) \textbf{episode constraint violation} (ECV) $\sum_t^{T_{\rm ep}} c_t$ shows the constraint satisfaction of intermediate polices and we do not terminate an episode when the constraint is violated; and (3) \textbf{cumulative training violations} (CTV) $\sum_t^{T} c_t$ indicates the exploration safety of algorithms, where $T$ is the total interactions since the training begins. CTV is the main criterion we focus on for safety-critical RL tasks. Note that all baselines except CPO and PPO-L are off-policy ones. Therefore, several transitions of a random policy are added to the buffer before the updates. That is why the training violation curves in some tasks and the total env interactions (x-axis) do not start from zero. Each run of one algorithm per task is repeated with 5 random seeds and we record the mean and the 95\% confidence interval.

The results are show in Fig~\ref{fig.comparison_curves}. As expected, the proposed DRPO converges to a policy with comparable or sometimes even better performance and nearly perfect constraint satisfaction across all tasks. Furthermore, DRPO outperforms all baselines w.r.t. CTV in all tasks, showing its safety during the whole training process. Among all tasks, model-based algorithms (DRPO, MBPO and SMBPO) have fewer CTVs due to their high sample efficiency. Besides the model rollouts, DRPO reduce violations further with the proposed DRC and the shield policy and we investigate the specific effects of each component in the ablation study. Therefore, the ECV of DRPO decreases faster than other baselines in most of the tasks and reaches a zero-violation policy earlier except in {Cartpole-Move}. We hypothesize that cartpole has simple dynamics so modeling the uncertainty may not be that significant for efficient learning. Moreover, due to the instablility of the system, little deviation of the policy outputs will cause the state to be unsafe because of high-risk-high-reward property of this case. One shortcoming of DRPO is that sometimes the ER is slightly lower than other baselines due to the shield policy considering uncertainty and the coarse line search. However, we believe it is acceptable with the presence of much fewer training violations.

\subsection{Ablation Study}

\begin{figure}[!bp]
\vspace{-5mm}
\centering
\subfloat[Quadrotor]{\label{fig:ablation_modules_quadrotor}\includegraphics[width=0.23\textwidth]{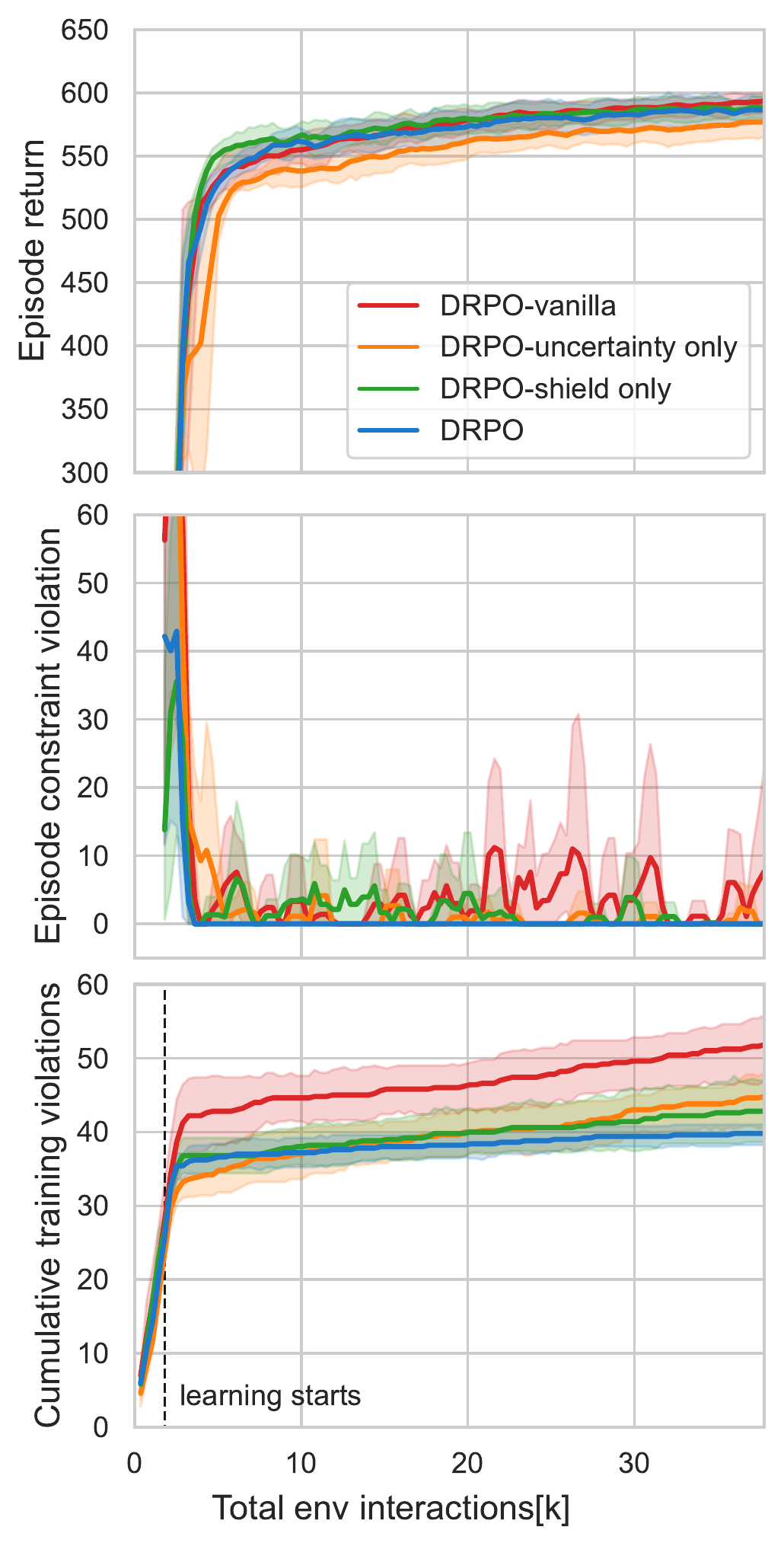}}\hfill
\subfloat[CarGoal]{\label{fig:ablation_modules_car}\includegraphics[width=0.23\textwidth]{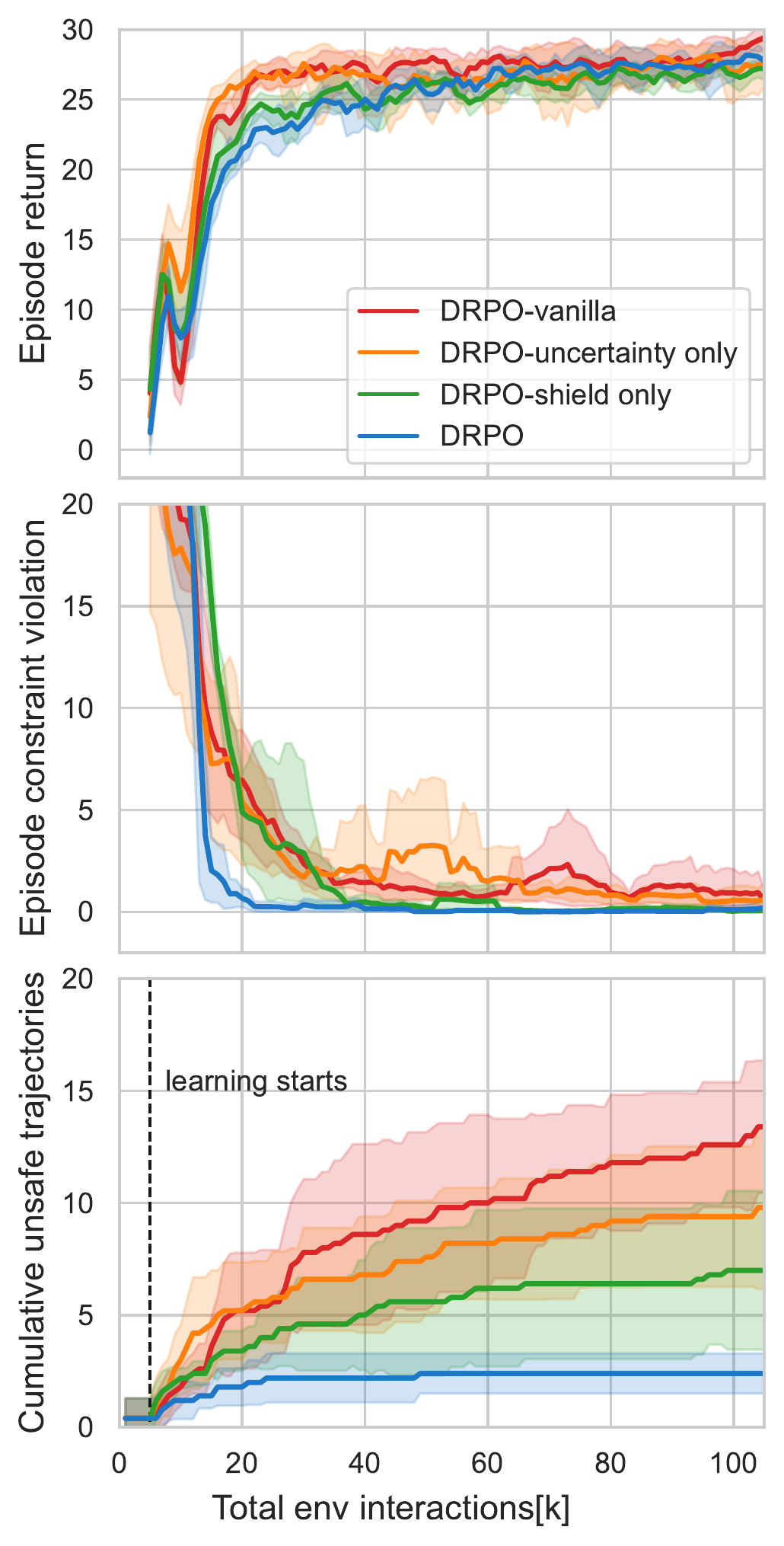}}
\caption{Training curves of DRPO and its ablation versions on Quadrotor (left column) and CarGoal (right column).}
\label{fig.ablation_modules}
\vspace{-3mm}
\end{figure}

\textbf{Ablating sub-modules}\quad To dig further about the effectiveness of the proposed modules, we ablate the DRC and shield policy respectively and conduct tests on {Quadrotor} and {CarGoal}. We denote \textit{DRPO-vanilla} as the one removing both DRC and shield policy, \textit{DRPO-uncertainty only} as the one removing shield policy, \textit{DRPO-shield only} as the one removing DRC and \textit{DRPO} as the proposed algorithm. Although DRPO-vanilla is strong enough, results shown in Fig~\ref{fig.ablation_modules} validate the advantages of both DRC and shield policy in terms of safety during training and evaluation at the cost of little loss in ER. Adding either DRC or the shield policy reduces CTVs, leading to better safety when they are combined in \textit{DRPO}. However, the evaluation safety of \textit{DRPO-uncertainty only} (mid row, orange curves) is not stable because it will violate constraints at times. We attribute this to the limited multipliers. Theoretically, $\lambda(s)$ should be $+\infty$ when $Z_\beta(s,a;\pi_h)>0$ but if $Z_\beta(s,a;\pi_h)\rightarrow0_+$ and $\lambda_{\rm max} Z_\beta(s,a;\pi_h)\approx0$, the policy will pay little attention to the constraint term in the Lagrangian, leading to unsafe actions. Fortunately, this can be rectified by the shield policy because it overwrite the action when $Z_\beta(s,a;\pi_h)>0$. One may question that DRPO is a simple combination of two methods (DRC and shield policy) but it is not the case. The $\beta$-shield policy comes exactly from minimization of the DRC and the formulation in~\eqref{eq.formal_drcrl_problem} bridges the certificate and the shield policy (mid part in Fig~\ref{fig.framework}). Moreover, the shield policy serves as a supplement for insufficient updates of the approximators of its source, i.e. the DRC and the corresponding constrained optimization problem. 

\begin{figure}[!h]
\vspace{-5mm}
\centering
\subfloat[Choosing different $\Phi^{-1}(\beta)$ means different confidence levels of uncertainty ]{\label{fig:ablation_2_beta}\includegraphics[width=0.23\textwidth]{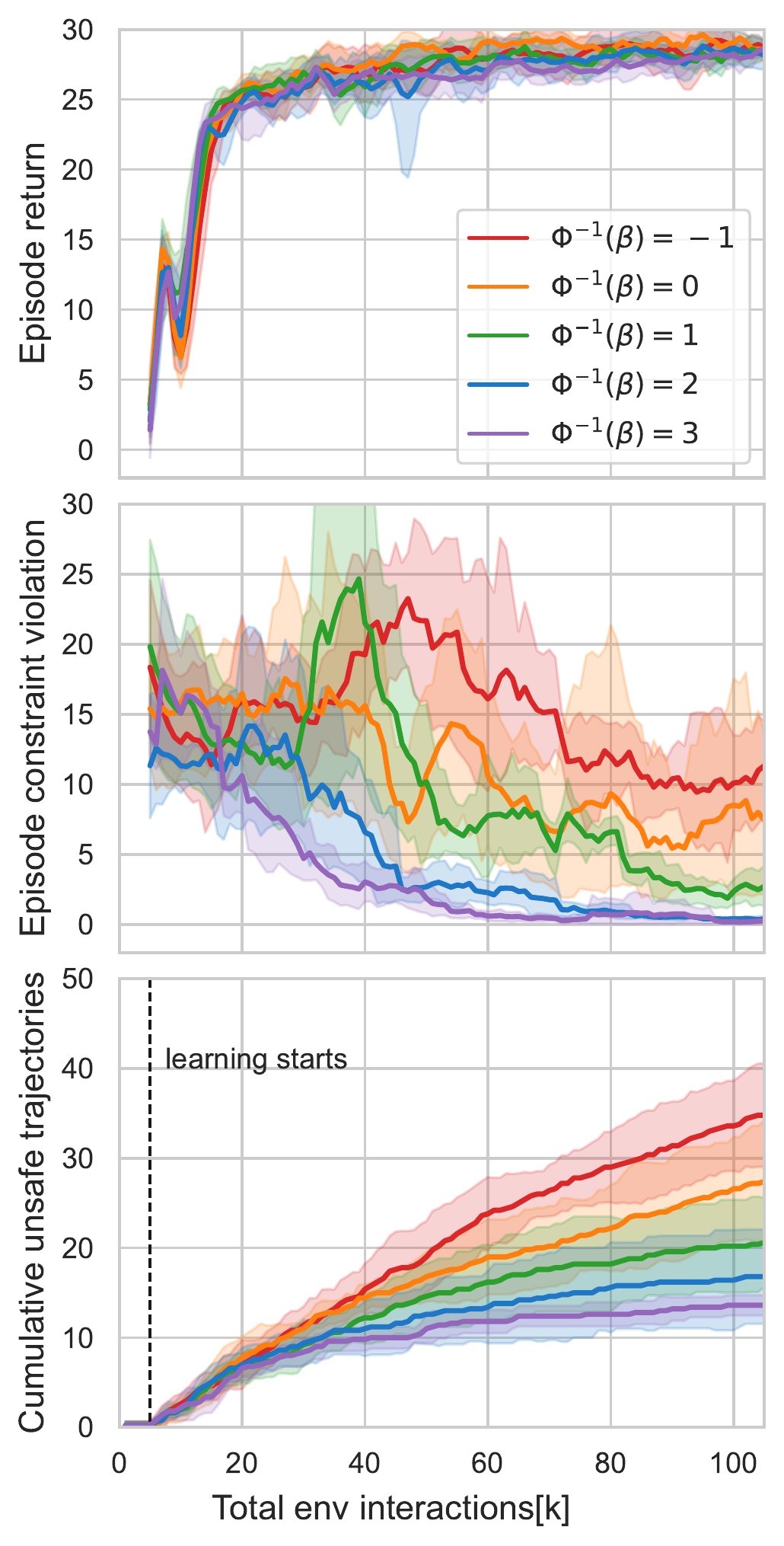}}\hspace{2mm}
\subfloat[Imposing constraints on current policy or shield policy]{\label{fig:ablation_3_constraints}\includegraphics[width=0.23\textwidth]{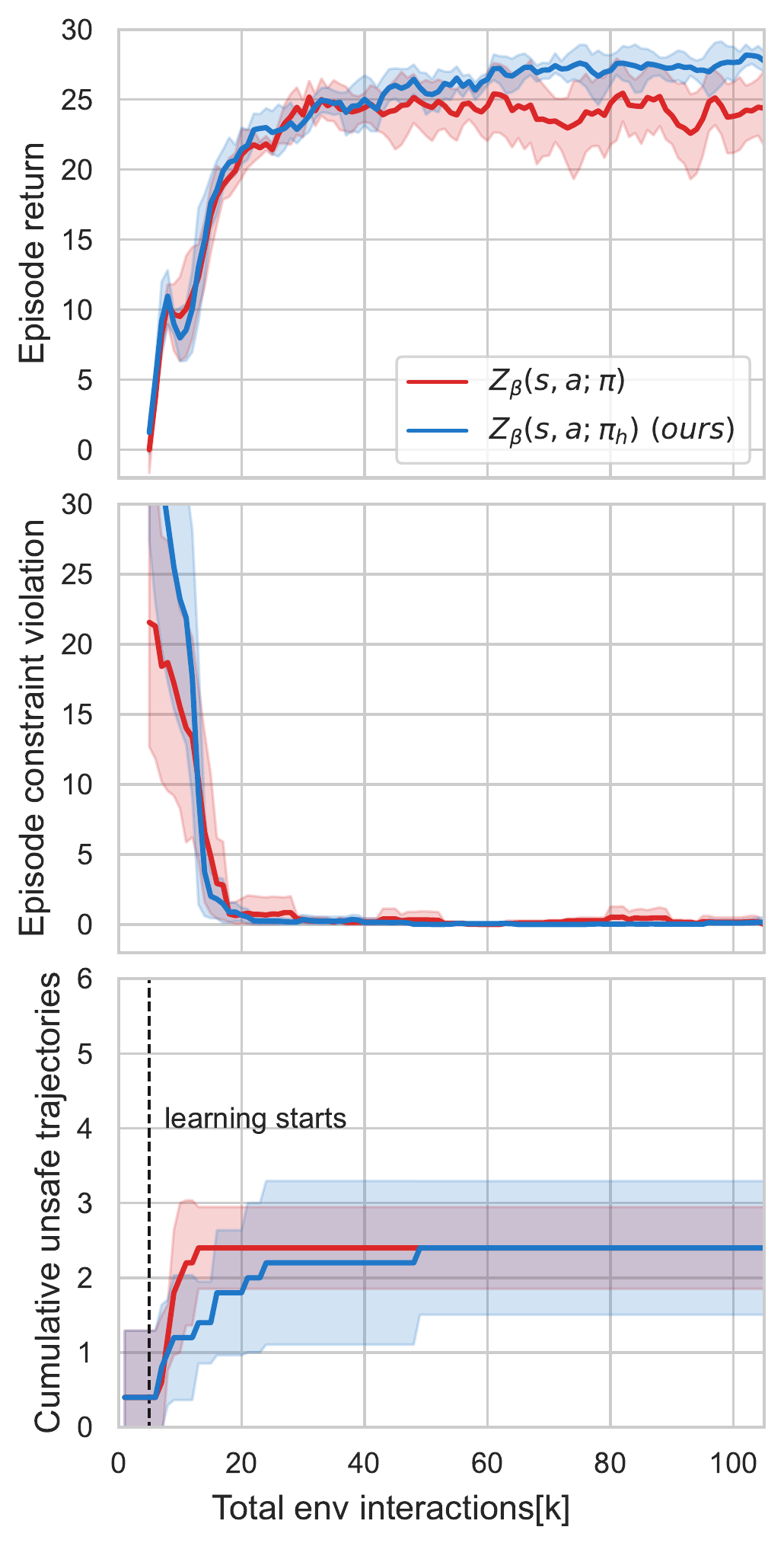}}
\caption{Training curves of DRPO and its ablation versions of different $\Phi^{-1}(\beta)$ (left column) and different constraint formulations (right column) on CarGoal.}
\label{fig.ablation_others}
\vspace{-1mm}
\end{figure}

\textbf{Choosing varying $\Phi^{-1}(\beta)$}\quad To study the effects of considering different levels of uncertainty, we compare DRPO with varying $\Phi^{-1}(\beta)$ from $-1$ to $3$ on CarGoal. Note that we disable the safety shield to focus on the uncertainty confidence level. In order to make collecting rewards conflict with staying safe, we do not terminate the trajectory when the constraint is violated so we calculate the unsafe trajectories rather than cumulative violations and the task becomes more difficult. Results in Fig~\ref{fig:ablation_2_beta} show that as $\Phi^{-1}(\beta)$ increases (more uncertainty is considered by the agent), the violations during training is becoming fewer and the final policy tends to be totally safe even without a shield. However, the ER only changes slightly when $\Phi^{-1}(\beta)$ changes and we attribute this to the decreasing model uncertainty. As the model learning proceeds, the variance of the predicted trajectory is getting lower and considering different levels of uncertainty has similar impacts. To summarize, choosing $\Phi^{-1}(\beta)=2$ (i.e., covering safety of 97\% of the possible subsequent cases) is enough for tasks in this work.


\textbf{Comparing constraint formulations}\quad To study whether imposing constraints on the agent to be inside the feasible set of \textit{a shield policy} will reduce conservativeness of the policy as expected in Section~\ref{sec.shield}, we compare DRPO with the baseline arising from~\eqref{eq.drcrl_problem}. The key difference lies in that the baseline constrains agents inside the feasible set of the \textit{current policy}. We also implement a shield as well as the line search method based on the current policy by performing several GD steps on the objective $\min_\pi Z_\beta(s,a;\pi)$ at each update step, which is similar to the policy improvement step in conventional RL but w.r.t. the safety specification. We denote the baseline as $Z_\beta(s,a;\pi)$ and DRPO as $Z_\beta(s,a;\pi_h)$ and the results are shown in Fig~\ref{fig:ablation_3_constraints}. Despite the similar safety performance (both ECVs and CTVs), the proposed constraint on $Z_\beta(s,a;\pi_h)$ does increases the cumulative rewards. This is because the initial problem restricts the agent inside the feasible set of an intermediate policy while the improved one encourages it to explore a larger feasible set $\SPACE{S}_\beta$. A larger workspace where the agent is free to explore means that the agent is able to approach safe states with possibly more rewards. Even if the danger is going to happen, the shield policy will intervene and override the unsafe actions, pulling the agent back to safe states. Thus, DRPO reaches higher return while maintains safe during training and evaluation.

\section{Conclusion}
In this paper, we propose DRPO, a safe RL algorithm, to address the model-uncertainty in model-based safe RL, thereby improving training time and evaluation safety. We first extend the reachability certificate to a distributional setting, realizing safety robust to model discrepancy or probabilistic prediction. The shield policy obtained from the DRC serves as the constraint in the problem formulation to further improve performance. The shield is leveraged in a switch-based scheme during training and a line search method during evaluation to maintain safety and reduce conservativeness. DRPO solves the constrained optimization problem with primal-dual method and deep neural network approximators. We evaluate DRPO and seven baselines on MuJoCo, safe-control-gym and safety-gym. Results show that DRPO outperforms all other algorithms in terms of policy safety while reaches competitive return performance. We also validate the effectiveness of the proposed DRC and the shield policy as well as the constraint formulation. Future work may include deploying DRPO on real-world robots to evaluate its practical ability. Moreover, if starting from a trivial but safe policy and a well-calibrated model, whether DRPO can be improved to realize provably safe learning remains an interesting question.




\section*{APPENDIX}
\subsection{Detailed Hyperparameters}
The hyperparameters of implementated algorithms are listed in Table~\ref{tab:off_hyper} and Table~\ref{tab:on_hyper} respectively.

\begin{table}[!htb]
\centering
\caption{Off-policy Algorithms Hyperparameters}
\label{tab:off_hyper}
\begin{tabular}{@{}ll@{}}
\toprule
Parameter                                  & Value                                    \\ \midrule
\textit{Shared}                            &                                          \\
\quad Optimizer                                   & Adam ($\beta_1=0.99, \beta_2=0.999$)     \\
\quad Approximation function                      & Multi-layer Perceptron (MLP)             \\
\quad \# of hidden layers                     & 2                                  \\
\quad \# of neurons per layer                 & 256                                \\
\quad Nonlinearity of hidden layer                & ReLU                               \\
\quad Critics learning rate (lr)                      & cos anneal (c.a.) 3e-4 $\rightarrow$ 8e-5 \\
\quad Actors lr                        & 
    \begin{tabular}[c]{@{}l@{}}
        Cartpole: c.a. 1e-4 $\rightarrow$ 4e-5 \\
        Quadrotor: c.a. 1e-4 $\rightarrow$ 4e-5 \\
        Safety-Gym: c.a. 8e-5 $\rightarrow$ 4e-5
    \end{tabular} \\
\quad Temperature factor $\alpha$ lr   & same as initial lr of actors                       \\
\quad Discount factor ($\gamma$)          & 0.99                                     \\
\quad \# of total episodes                       &                           
    \begin{tabular}[c]{@{}l@{}}
        Cartpole: 50 \\
        Quadrotor, Safety-Gym: 100 \\
    \end{tabular}\\
\quad \# of critics updates per step             & 10                                \\
\quad \# of actors updates per step              & 5                                        \\
\quad \# of multiplier updates per step          & 2                                        \\
\quad Target smoothing coefficient ($\tau$)      & 0.005                                    \\
\quad Expected entropy ($\bar{\mathcal{H}}$)     & -$\mathrm{dim}(\SPACE{A})$               \\
\quad Replay buffer size                         & 
    \begin{tabular}[c]{@{}l@{}}
        Cartpole, Safety-Gym: 500k \\
        Quadrotor: 360k
    \end{tabular} \\
\quad Replay batch size                          & 256                         \\
\quad Buffer warm-up size                        & 
    \begin{tabular}[c]{@{}l@{}}
        Cartpole: 1,000 \\
        Safety-Gym: 5,000 \\
        Quadrotor: 1,800
    \end{tabular} \\
\quad Mix-up ratio                           & 10\% real data, 90\% virtual data   \\
\quad Model approximator                         & 5 MLPs                      \\
\quad \# of model hidden layers              & 2 trunk, 1 for $\mu_\phi$ and $\sigma_\phi$ head \\
\quad \# of neurons per model layer          & 200                                \\
\quad Model lr                                   & 1e-3                      \\
\quad \# of model initial updates                                   & 10,000                  \\
\quad \# of model updates interval                                 & $T_\mathrm{ep}/4$        \\
\quad \# of GD steps per model update                               & 1,000                    \\
\quad Nonlinearity of model hidden layer         & swish                               \\
\quad Rollout batch size                         & 100                         \\
\quad Rollout length                             & 10                          \\
\midrule
\textit{DRPO, RAC}                             &                                          \\
\quad Multiplier lr                            & c.a. 3e-4 $\rightarrow$ 1e-5 \\ 
\midrule
\textit{CSC, SAC-L}                              &                                          \\
\quad Multiplier lr                              & 3e-4                                   \\
\quad Critic conservative coefficient               & 0.05                                   \\
\midrule
\textit{SMBPO}                                   &                                          \\
\quad Look-ahead horizon                         & 10 \\
\bottomrule
\end{tabular}
\end{table}

\begin{table}[!htb]
\centering
\caption{On-policy Algorithms Hyperparameters}
\label{tab:on_hyper}
	\begin{tabular}{@{}ll@{}}
		\toprule
		Parameter & Value \\
		\hline
		\emph{Shared} & \\
		\quad Optimizer                     &  Adam ($\beta_{1}=0.9, \beta_{2}=0.999$)\\
		\quad Approximation function        & MLP \\
		\quad \# of hidden layers           & 2\\
		\quad \# of neurons per layer       & 256\\
		\quad Nonlinearity of hidden layer  & Tanh\\
		\quad Critic learning rate (lr)     & 1e-3\\
		\quad Actor lr                      &  3e-4\\
		\quad Discount factor ($\gamma$)    & 0.99\\
		\quad GAE parameter ($\lambda$)     &  0.97\\
		\quad Batch size                    &  4000\\
		\quad \# of actors                  & 4\\
		\quad \# of critic updates per iteration  & 80\\
		\quad Target KL                     &  0.01\\
		\quad Cost threshold                & 0\\
		\midrule
		\emph{PPO-Lagrangian} &\\
		\quad Clip ratio                    &  0.2\\
		\quad Max \# of actor updates per iteration  & 80\\
		\quad KL margin                     &  1.2\\
		\quad Initial multiplier            &  1.0\\
		\quad Multiplier lr                 &  5e-2 \\
		\quad \# of multiplier updates per iteration  & 1\\
		\midrule
		\emph{CPO} &\\
		\quad Backtrack coefficient         & 0.8\\
		\quad Max \# of backtrack iterations          & 10\\
		\bottomrule
	\end{tabular}
\end{table}


\bibliography{IEEEabrv, ref}
\bibliographystyle{IEEEtran}


 

\vspace{-8mm}
\begin{IEEEbiography}
[{\includegraphics[width=1in,height=1.25in,clip,keepaspectratio]{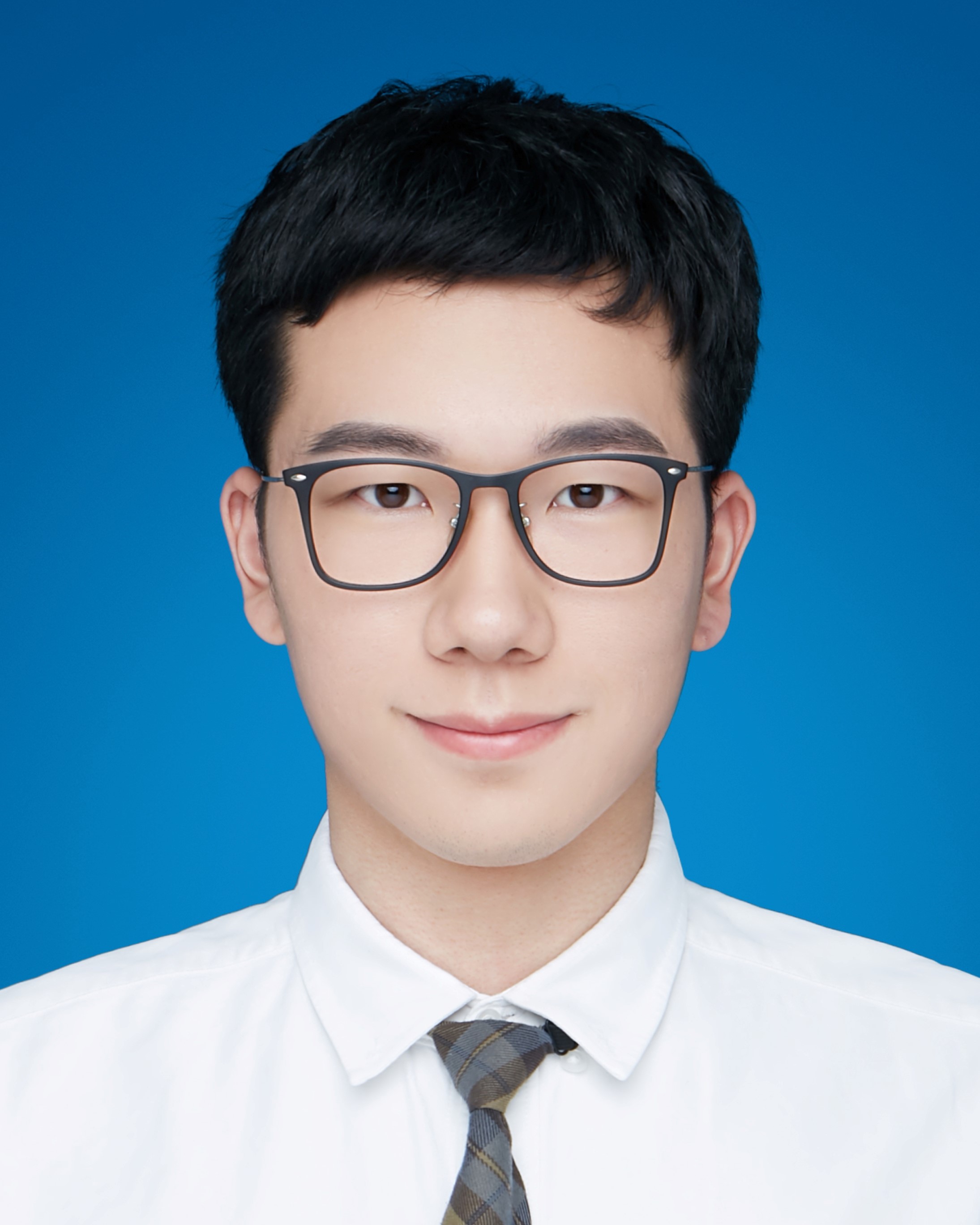}}]{Dongjie Yu} received the B.S. degree in automotive engineering from Tsinghua University, Beijing, China, in 2020, where he is currently pursuing the M.S. degree in mechanical engineering.

His current research interests include safe reinforcement learning and its application in the decision-making and control of robotics and autonomous driving.
\end{IEEEbiography}

\vspace{-4mm}

\begin{IEEEbiography}
[{\includegraphics[width=1in,height=1.25in,clip,keepaspectratio]{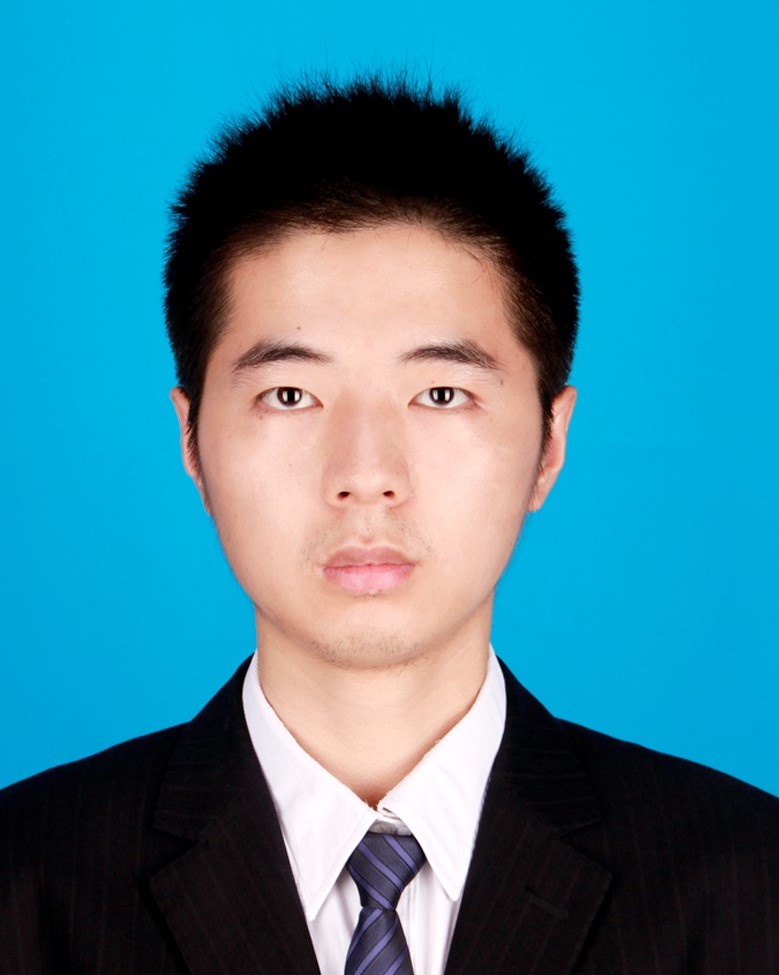}}]{Wenjun Zou} received the B.S. degree in automotive engineering from Tsinghua University, Beijing, China, in 2020. He is currently pursuing the Ph.D. degree in mechanical engineering from Tsinghua University, Beijing, China.

His current research interests include decision and control of autonomous vehicles and reinforcement learning.
\end{IEEEbiography}

\vspace{-4mm}

\begin{IEEEbiography}
[{\includegraphics[width=1in,height=1.25in,clip,keepaspectratio]{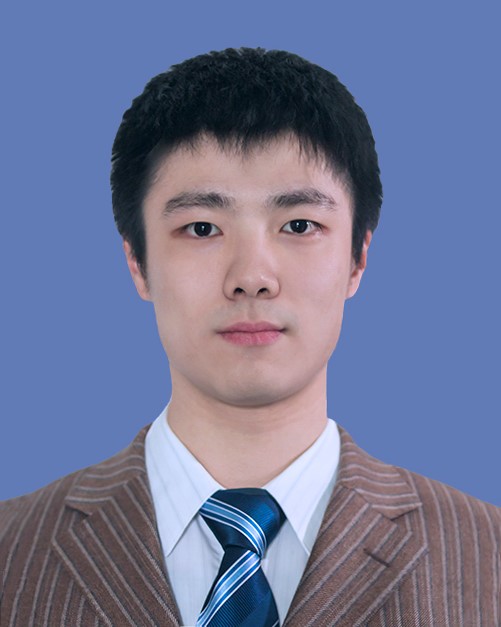}}]{Yujie Yang} received the B.E. degree in automotive engineering from Tsinghua University, Beijing, China, in 2021. He is currently pursuing his Ph.D. degree in the School of Vehicle and Mobility, Tsinghua University, Beijing, China.

His research interests include decision and control of autonomous vehicles, reinforcement learning, and optimal control.
\end{IEEEbiography}
\vspace{-3mm}
\begin{IEEEbiography}
[{\includegraphics[width=1in,height=1.25in,clip,keepaspectratio]{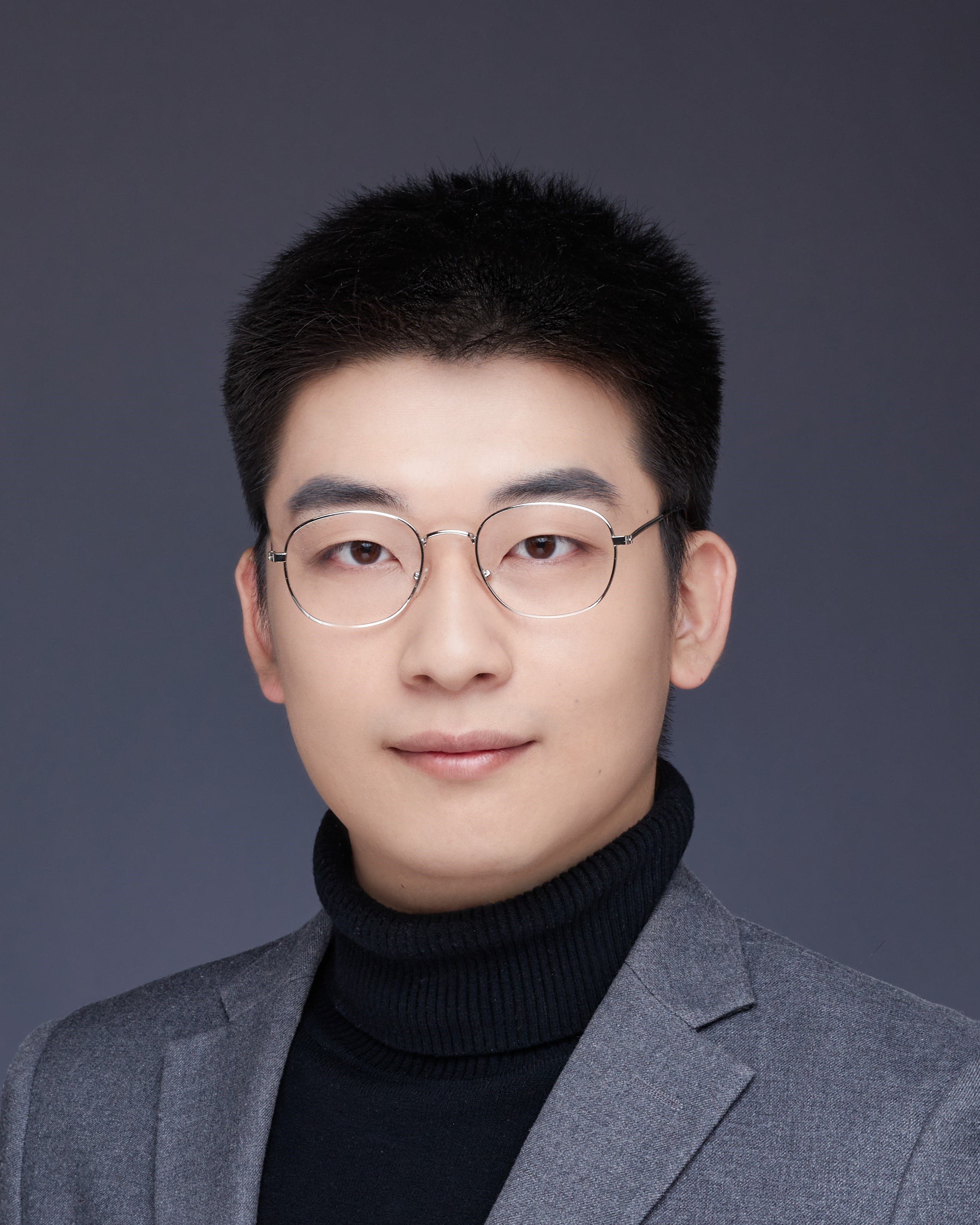}}]{Haitong Ma}received the M.S. and B.S. degree in vehicle engineering from Tsinghua University in 2022, 2019, respectively. Now he is a first-year Ph.D. student at Harvard John A. Paulson School of Engineering and Applied Sciences (SEAS).

His research interest lies in the intersection of control theory and machine learning. He received the outstanding master graduate and master thesis of Tsinghua University, L4DC best paper award finalists, ITSC best student paper award, championship of Honda Eco Mileage challenge in China, etc.
\end{IEEEbiography}
\vspace{-3mm}
\begin{IEEEbiography}
[{\includegraphics[width=1in,height=1.25in,clip,keepaspectratio]{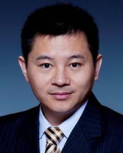}}]{Shengbo Eben Li}(Senior Member, IEEE) received the M.S. and Ph.D. degrees from Tsinghua University in 2006 and 2009, respectively.

He worked at Stanford University, the University of Michigan, and the University of California at Berkeley. He is currently a Tenured Professor at Tsinghua University. He is the author of over 100 journals/conference papers, and the co-inventor of over 20 Chinese patents. His research interests include intelligent vehicles and driver assistance, reinforcement learning and distributed control, and optimal control and estimation. He was a recipient of the Best Paper Award in 2014 IEEE ITS Symposium, the Best Paper Award in 14th ITS Asia Pacific Forum, the National Award for Technological Invention in China in 2013, the Excellent Young Scholar of NSF China in 2016, and the Young Professorship of Changjiang Scholar Program in 2016. He serves as an Associate Editor for \textit{IEEE Intelligent Transportation Systems Magazine} and the \textsc{IEEE Transactions on Intelligent Transportation Systems}.
\end{IEEEbiography}
\vspace{-3mm}
\begin{IEEEbiography}
[{\includegraphics[width=1in,height=1.25in,clip,keepaspectratio]{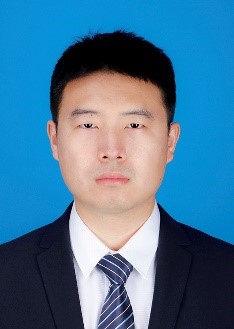}}]{Jingliang Duan}received his Ph.D. degree in the School of Vehicle and Mobility, Tsinghua University, China, in 2021. He studied as a visiting student researcher in the Department of Mechanical Engineering, University of California, Berkeley, in 2019, and worked as a research fellow in the Department of Electrical and Computer Engineering, National University of Singapore, from 2021 to 2022. He is currently an associate professor in the School of Mechanical Engineering, University of Science and Technology Beijing, China.

His research interests include reinforcement learning, optimal control, and self-driving decision-making..
\end{IEEEbiography}
\vspace{-3mm}
\begin{IEEEbiography}
[{\includegraphics[width=1in,height=1.25in,clip,keepaspectratio]{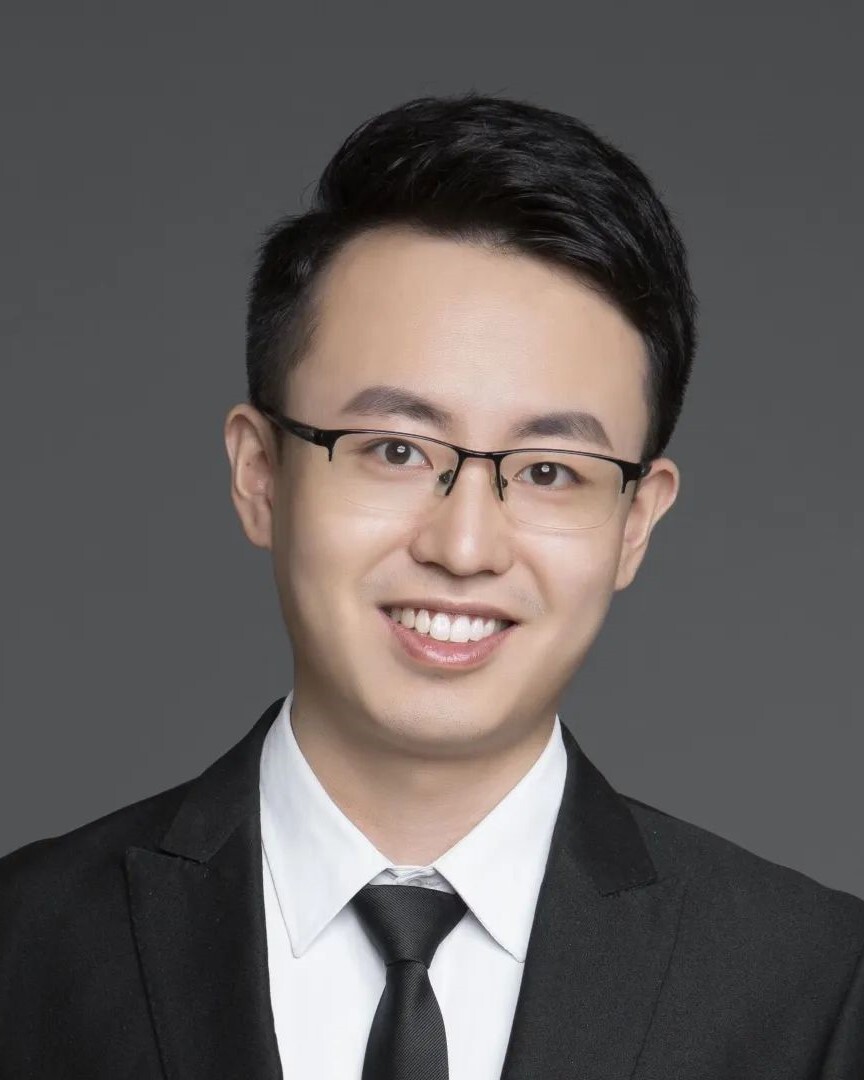}}]{Jianyu Chen}is an assistant professor in Institute for Interdisciplinary Information Sciences (IIIS) at Tsinghua University starting from 2020. Prior to that, he was working with Prof. Masayoshi Tomizuka at the University of California, Berkeley and received his Ph.D. degree in 2020. He received his Bachelor degree from Tsinghua University in 2015. He is working in the cross fields of robotics, reinforcement learning, control and autonomous driving. His research goal is to build advanced robotic systems with high performance and high intelligence.
\end{IEEEbiography}


\vfill

\end{document}